\pdfoutput=1

\documentclass[11pt]{article}

\usepackage{acl}

\usepackage{times}
\usepackage{latexsym}

\usepackage{bm}
\usepackage{amsfonts}
\usepackage{amsmath}
\usepackage{mathrsfs}
\usepackage{graphicx}
\usepackage{booktabs}
\usepackage{multirow}
\usepackage{float}
\usepackage{subcaption}
\usepackage{stfloats}
\usepackage{amssymb}
\usepackage{xparse}
\usepackage{booktabs}
\usepackage{colortbl}
\usepackage{xspace}
\usepackage{enumitem}
\usepackage{todonotes}
\usepackage{bbm}

\usepackage{tcolorbox}
\newcounter{mybox}

\usepackage{soul} 
\usepackage{xcolor} 

\definecolor{lightblue}{RGB}{236,244,255}

\newcommand{\highlight}[1]{\sethlcolor{lightblue}\hl{#1}}

\usepackage{array}

\usepackage[normalem]{ulem}
\useunder{\uline}{\ul}{}

\usepackage{pifont}

\usepackage[T5]{fontenc}

\usepackage[utf8]{inputenc}

\usepackage{microtype}

\usepackage{inconsolata}

\usepackage{graphicx}

\title{Speech is More Than Words: \\ Do Speech-to-Text Translation Systems Leverage Prosody?}
% ...

\author{Ioannis Tsiamas$^{\diamondsuit}$\thanks{\, \, Work done during an internship at Apple.}\quad Matthias Sperber${^\dagger}$\quad  Andrew Finch${^\dagger}$$\quad$ Sarthak Garg${^\dagger}$ \\ 
$^{\diamondsuit}$Universitat Politècnica de Catalunya $\quad$
${^\dagger}$Apple\\
\normalsize{\texttt{ioannis.tsiamas@upc.edu, sperber@apple.com}}
}

\newcommand{\benchmarkname}{\textsc{ContraProST}}
\newcommand{\mt}{\text{MT}}
\newcommand{\asr}{\text{ASR}}
\newcommand{\stt}{\text{S2TT}}
\newcommand{\sts}{\text{S2ST}}
\newcommand{\strong}{C_\mathcal{G}}
\newcommand{\weak}{C_\mathcal{D}}

\newcommand{\citenllb}{\hyperlink{cite.nllb}{(NLLB Team, 2022)}}
\newcommand{\citeSeamless}{\hyperlink{cite.seamless}{(Seamless Communication, 2023a)}}
\newcommand{\citeSeamlessv}{\hyperlink{cite.seamlessv}{(Seamless Communication, 2023b)}}
\newcommand{\citegpt}{\hyperlink{cite.gpt}{(OpenAI, 2024)}}

\begin{document}
\maketitle

\begin{abstract}

The prosody of a spoken utterance, including features like stress, intonation and rhythm, can significantly affect the underlying semantics, and as a consequence can also affect its textual translation. Nevertheless, prosody is rarely studied within the context of speech-to-text translation (\stt) systems. In particular, end-to-end (E2E) systems have been proposed as well-suited for prosody-aware translation because they have direct access to the speech signal when making translation decisions, but the understanding of whether this is successful in practice is still limited. A main challenge is the difficulty of evaluating prosody awareness in translation. To address this challenge, we introduce an evaluation methodology and a focused benchmark (named \benchmarkname{}) aimed at capturing a wide range of prosodic phenomena. Our methodology uses large language models and controllable text-to-speech (TTS) to generate contrastive examples. Through experiments in translating English speech into German, Spanish, and Japanese, we find that (a)~\stt{} models possess some internal representation of prosody, but the prosody signal is often not strong enough to affect the translations, (b)~E2E systems outperform cascades of speech recognition and text translation systems, confirming their theoretical advantage in this regard, and (c)~certain cascaded systems also capture prosodic information in the translation, but only to a lesser extent that depends on the particulars of the transcript's surface form.\footnote{\href{https://github.com/apple/ml-speech-is-more-than-words}{github.com/apple/ml-speech-is-more-than-words}}
\end{abstract}

\section{Introduction}

\begin{table}[h!]
\centering
\resizebox{1\columnwidth}{!}{
    \begin{tabular}{@{}lll@{}}
        \toprule \toprule
        \multicolumn{3}{c}{Example: \emph{These are German teachers.}} \\ 
        \midrule
        \multirow{3}{*}{{\color[HTML]{F56B00} A}} & Prosody & These are {\color[HTML]{F56B00} GERMAN} teachers. \\
                           & Explanation & Teachers from Germany \\
                           & Translation & Dies sind {\color[HTML]{F56B00} Deutschlehrer}. \\ 
        \midrule
        \multirow{3}{*}{{\color[HTML]{3166FF} B}} & Prosody & These are German {\color[HTML]{3166FF} TEACHERS}. \\
                           & Explanation & Teachers that teach German \\
                           & Translation & Dies sind {\color[HTML]{3166FF} deutsche Lehrer}. \\ 
        \midrule \midrule
        \multicolumn{3}{c}{Example: \emph{John laughed at the Party.}} \\ 
        \midrule
        \multirow{3}{*}{{\color[HTML]{F56B00} A}} & Prosody & John {\color[HTML]{F56B00} LAUGHED (pause)} at the Party. \\
                           & Explanation & Laughed while at the party (literal) \\
                           & Translation & John lachte {\color[HTML]{F56B00} während} der Party. \\ 
        \midrule
        \multirow{3}{*}{{\color[HTML]{3166FF} B}} & Prosody & John {\color[HTML]{3166FF} LAUGHED AT (pause)} the Party. \\
                           & Explanation & Ridiculed the party (idiomatic) \\
                           & Translation & John lachte {\color[HTML]{3166FF} über} die Party. \\ 
        \bottomrule \bottomrule
    \end{tabular}
}
\caption{Examples of prosody-aware Speech Translation from English to German.}
\label{tab:intro_examples}
\end{table}

Prosody, which includes features like stress, intonation, and rhythm, is crucial for conveying meaning in spoken language beyond the literal words used~\cite{ladd1980,Bolinger1989}. Among others, prosody can direct focus and clarify meaning~\cite{Bolinger1961,Halliday1967}, disambiguate syntax and sentence structure~\cite{Bolinger1989}, convey the emotional state of the speaker~\cite{Banse1996}, and provide useful cues that make communication more effective~\cite{Shriberg1998}. For example, the phrase ``\emph{Really?}'' can express surprise, genuine interest or disbelief, depending on the intonation with which is spoken.

Table~\ref{tab:intro_examples} illustrates the importance of considering prosody when generating translations in \stt. 
~\citet{taking_stock} suggest that E2E \stt{} systems may have an inherent advantage over cascaded systems in this regard, because only the former have access to the speech signal when making translation decisions. However, our understanding of whether prosody informs translation choices in practice is currently still limited, as prior research on this topic either shows only anecdotal evidence~\cite{st_industrial_practice}, focuses on only a small subset of prosodic phenomena ~\cite{prosody_st_korean,meld_st}, or considers how prosody informs target-side speech with regards to generated prosody but not lexical choice (\S\ref{sec: related_work}).

In this paper, we take steps toward a reliable and comprehensive evaluation methodology, which is one of the most important prerequisites for achieving prosody-aware \stt{}. We identify three central challenges that must be addressed: (1)~Existing \stt{} benchmarks often do not include prosody-rich spontaneous speech and/or do not include translations that are informed by the audio, limiting the extent to which reference translations are influenced by source-side prosody. (2)~General-purpose evaluation methods like \textsc{BLEU}~\cite{bleu} and \textsc{COMET}~\cite{xcomet} are insensitive to the often subtle changes in translation caused by input prosody. (3)~Existing prosody-centric benchmarks are difficult to scale to broader coverage of languages and prosodic phenomena, which hinders comprehensive analysis.

To address these challenges, we take inspiration from prior work on behavioral testing~\cite{ribeiro-etal-2020-beyond,behavioral_mt} and contrastive evaluation~\cite{contrastive_eval1}. We address the first challenge by synthesizing prosody-rich data that covers a wide range of prosodic phenomena through the use of large language models (LLMs) and controllable TTS (cTTS). We tackle the second challenge by developing a double-contrastive evaluation approach, i.e.\ a directional behavioral test that relies on minimal pairs (differing only in prosody) to evaluate prosody-awareness in \stt{} in isolation. The resulting benchmark, \benchmarkname{} (\underline{Contra}stive \underline{Pro}sody \underline{ST}), covers a variety of language pairs and prosodic phenomena. Since it is mostly automated, it can be further extended, thus addressing also the third challenge.

To investigate how well current state-of-the-art models understand and leverage prosody, 
we evaluate \stt{} models of various sizes and types, including both E2E and cascaded systems. We find indications that \stt{} models represent prosody internally, but this knowledge is often not manifested in the translations. We observe that while tested cascaded systems perform better on traditional evaluation (\textsc{COMET}), E2E models outperform cascaded models on \benchmarkname{}. We also find indications that some amount of prosody is carried through transcripts in cascaded setups, but this depends on the particulars of the transcriptions. The most important implication of our findings is the need for exploring improvements of \stt{} regarding prosody-awareness, e.g.\ through auxiliary losses or finetuning on prosody-rich data.

\section{The \benchmarkname{} Benchmark} \label{sec:data_gen}

    \begin{figure*}[h]
        \centering
        \includegraphics[width=0.9\linewidth]{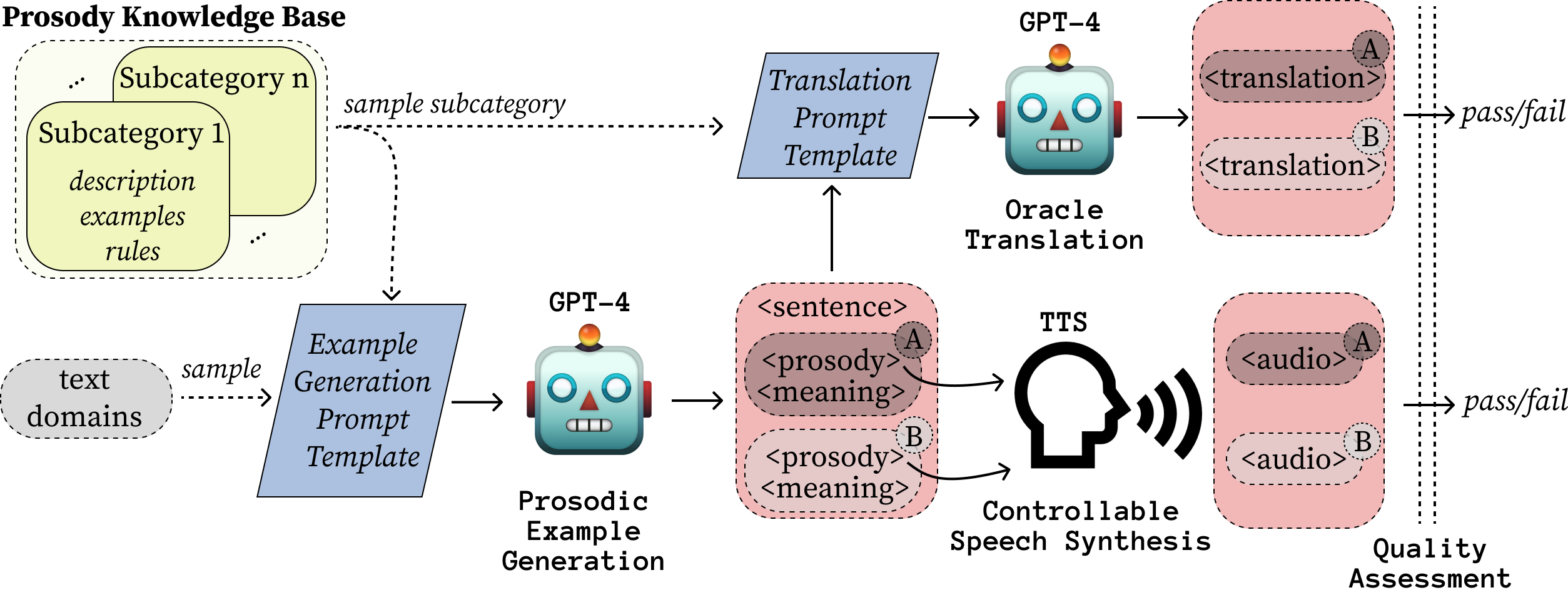}
        \caption{The Data Generation process for \benchmarkname{}.}
        \label{fig:data_generation}
    \end{figure*}

    \benchmarkname{} is composed of double-contrastive examples (see Table~\ref{tab:intro_examples}), where each example is composed of a sentence in English that could be semantically ambiguous, along with two different pairs of $<$speech, translation$>$ that capture contrastive cases of prosody.

    As it would be expensive and practically difficult to collect such test data manually, we employ an automatic data generation process, illustrated in  Fig.~\ref{fig:data_generation}. First, we identify several relevant categories where prosody influences sentence semantics in important ways, and construct illustrative examples that reflect the respective phenomena of each category, while highlighting differences in prosody-induced meaning (\S\ref{subsec:categories}). We then prompt GPT-4\footnote{\textsc{GPT-4o-2024-05-13}}~\citegpt{} to generate sentences similar to the examples for each subcategory using in-context learning, grounding the generation on different text domains to increase diversity (\S\ref{subsec:prosodic_example_generation}). Next, GPT-4 is prompted to translate each prosodic case, while also being given access to the prosodies, meanings and general information of the category, thus acting as a prosody- and context-aware oracle translator (\S\ref{subsec:oracle_translation}). Finally, we use the OpenAI TTS API\footnote{\textsc{TTS-1-hd}, \href{https://platform.openai.com/docs/models/tts}{platform.openai.com/docs/models/tts}} to synthesize the prosodic speech of each case (\S\ref{subsec:controllable_speech_synthesis}). Each generation stage is coupled with filtering and quality assessment to ensure the data are of high quality.

    \subsection{Categorization of Prosodic Phenomena} \label{subsec:categories}

        Below, we summarize the examined prosodic categories. Details and examples are available in the Appendices \ref{app:subcategories} and \ref{app:examples}.

        \setlength{\parindent}{0pt}
    
        \textbf{(1) Sentence Stress.} This is usually manifested through increased loudness, vowel length or higher pitch~\cite{Fry1955}, invoking emphasis on certain words within a sentence, potentially changing the semantics by shifting focus~\cite{Wagner2020}. We further categorize prosodic stress in four subcategories according to the purpose of the stress or its use in disambiguation of linguistic phenomena (see Appendix~\ref{app:subcategories_sentence_stress}).\\
        \textbf{(2) Prosodic Breaks.} Here we consider the existence or placement of longer breaks in the flow of speech, primarily associated with tempo, that create different phrasal boundaries and help disambiguate syntax and sentence structure~\cite{Bolinger1989}. We follow~\citet{Hirschberg1992} and use the subcategories outlined in Appendix~\ref{app:subcategories_prosodic_breaks}. \\
        \textbf{(3) Intonation Patterns.} This concerns the modality of the sentence, specifically whether it is a statement (falling tone), or a declarative question (rising tone)~\cite{declarative_questions}. \\
        \textbf{(4) Emotional Prosody.} A different emotional tone can indicate a speaker's emotional state and thus affect the semantics of the utterance~\cite{Banse1996}. Emotional tone is usually manifested through changes in pitch, tempo, and loudness. For example, happiness is associated with higher values in pitch and tempo, while sadness exhibits lower values for pitch, tempo, and loudness~\cite{sound_of_emotional_prosody}. Here, we focus on the seven \emph{basic} emotions: happy, sad, angry, disgust, surprisal, fear, and neutral~\cite{Ekman1971,Ekman1992}, based on which we construct all possible pairs, thus having 21 subcategories. \\
        \textbf{(5) Politeness.} The level of politeness can be conveyed by non-verbal cues, and influences the pragmatic context of a conversation. A polite tone is associated with a higher pitch and a smooth rhythm, while an impolite tone is manifested through low pitch, irregular rhythm and very high or low loudness levels~\cite{Culpeper2003,Culpeper2011}.
        \setlength{\parindent}{12pt}

        \begin{tcolorbox}[float, colback=gray!10, colframe=gray!80, title=Prompt 1: Prosodic Example Generation, label=box:prompt_example_generation, boxsep=1mm, left=2mm, right=2mm]
        \refstepcounter{mybox} \label{box:prompt_example_generation}
        \fontsize{10.5pt}{12pt}\selectfont
            You are a helpful assistant with expert knowledge in linguistics, speech, and prosody. Your task is to come up with examples of English sentences where different prosody would change the meaning of the sentence significantly.$^{(1)}$
            
            \highlight{\{Details for Category \& Subcategory\}}$^{(2)}$
            
            Here are some examples to guide you:
            
            \highlight{\{List of Examples\}}$^{(3)}$
            
            Strictly follow these rules:
            
            \highlight{\{List of Rules\}}$^{(4)}$
            
            Provide a rating of how significant is the difference between the two meanings.$^{(5)}$
            
            Generate \highlight{\{n\}} such examples, with rating as high as possible,$^{(6)}$ in the domain of \highlight{\{domain\}}.$^{(7)}$
        
        \end{tcolorbox}

    \subsection{Prosodic Example Generation} \label{subsec:prosodic_example_generation}

        For each category, we prompt GPT-4 to generate sentences based on hand-crafted category-specific examples. More specifically, we have the LLM generate English sentences, each with two different textual prosodic annotations and  respective meanings/interpretations to guide subsequent translation (\S\ref{subsec:oracle_translation}). The generated annotations include rich text that indicates different levels of emphasis, pause tags, and special punctuation such as ellipsis, exclamation, or interobang~(!?). The sentence itself is generated to be as simple as possible, ending with a full stop or question mark.

        The general prompt template is displayed in Prompt~\ref{box:prompt_example_generation}. It starts with some general information about the task, see superscript (1). The prompt then continues with details describing the current category/subcategory (2). The next part refers to in-context learning~\cite{gpt3}, where we provide a list of illustrative, hand-crafted examples for the LLM to follow~(3). In certain subcategories, due to repeated mistakes observed in preliminary explorations, we also provide examples to avoid. In~(4) we provide a list of rules for the LLM to adhere to, indicating the desired structure of the sentence and how to use prosodic notation, which might not be obvious from the examples~(3). Examples of such rules are ``do not include prosodic annotations in the sentence,'' or ``stress different noun-phrases in each prosodic case.'' We furthermore use \textit{self-criticism}~\cite{self-critic} by instructing the model to rate its own generations, according to how different the two prosodic interpretations are (5). Then we instruct the LLM to generate examples that have high scores after self-reflection (6). These scores are also used later during filtering. Finally, to avoid repetitive examples and enhance diversity, we condition the generation on specific text domains~(7)~\cite{text_domain}. The list of domains is also generated by GPT-4 based on the context that its subcategory would naturally occur (e.g.\ \textit{legal testimonies}). For each text domain in the subcategory the LLM then generates $n$ candidate examples. We use several hand-crafted text-based filtering steps to ensure that the examples generated by the LLM at this stage comply with the instructions specified in (4).

    \subsection{Oracle Translation} \label{subsec:oracle_translation}

        
        Recent research on the emerging capabilities of LLM-based \mt{}~\cite{palm2023_llm_mt,Alves2023_llm_mt,Zhang2023_llm_mt} has shown that LLMs can attain very high translation quality, especially for high-resource languages~\cite{Robinson2023_llm_mt_high_resource} and including translation factors such as emotions~\cite{brazier-rouas-2024-conditioning}, suggesting the possibility that LLMs can be leveraged for prosodic translation synthesis. To obtain the translations of the prosodic cases, we thus utilize GPT-4 as a prosody- and context-aware oracle translator. The LLM is prompted to translate, while having access to the sentence, the textual prosodic annotations (prosody-awareness), and the semantic interpretations (context-awareness). The template prompt is shown in Prompt~\ref{box:prompt_translation}. We provide a list of contraints to the LLM with several goals in mind: (i) avoid generating prosodic annotations in the translations; (ii) avoid translating the interpretations rather than the sentences; (iii) encourage the model to generate different translations for each case; (iv) ensure that differences in the translations are only due to the difference in the prosodies.

        \begin{tcolorbox}[float, colback=gray!10, colframe=gray!80, title=Prompt 2: Oracle Translation, label=box:prompt_translation, boxsep=1mm, left=2mm, right=2mm]
        \refstepcounter{mybox} \label{box:prompt_translation}
        \fontsize{10.5pt}{12pt}\selectfont
            You are a helpful assistant with expert knowledge in speech, prosody, linguistics and translation, particularly in English and \highlight{\{Target Lang\}}.
            You will be provided with a sentence in English (S) and two different prosodic variations (S$_\text{A}$, S$_\text{B}$), focused on \highlight{\{Category\}}, which correspond to two different semantic interpretations.
            
            Your task is to translate S, S$_\text{A}$ and S$_\text{B}$ into \highlight{\{Target Lang\}}, as T, T$_\text{A}$, and T$_\text{B}$.

            Carry out the translation in these steps:

            (1) Translate S into T.
            
            (2) Translate S$_\text{A}$ to T$_\text{A}$ and S$_\text{B}$ to T$_\text{B}$, by focusing on how T should change in order to reflect the additional information from the prosodies.
            
            The following constraints should be applied: \highlight{\{List of Constraints\}}

            The sentence S is: \highlight{\{sentence\}}
            
            The two different prosodic variations are:
            
            S$_\text{A}$. \highlight{\{prosody$_\text{A}$\}} (\highlight{\{meaning$_\text{A}$\}})
            
            S$_\text{B}$. \highlight{\{prosody$_\text{B}$\}} (\highlight{\{meaning$_\text{B}$\}})
        
        \end{tcolorbox}

        Although prosody variants substantially influence sentence semantics, this does not always imply that the ideal translations must differ. In particular, sometimes a translation that leaves semantics ambiguous may be preferred as the most natural translation.\footnote{This is essentially an instance of the fluency-accuracy trade-off \cite{lim-etal-2024-simpsons}.} As a consequence, constraint (iii) is sometimes overly strict and even in conflict with constraint (iv), leading to changes in the translations that do not stem from the prosodies, that are not idiomatic. To account for that, we include a post-editing step, where GPT-4 is instructed to choose the most fitting translation among $\{T, T_A, T_B\}$ for each prosodic case, independently from the other prosodic cases, while having access only the prosody information (Prompt~\ref{box:prompt_post_edit}). We prompt the LLM to first provide an explanation, before selecting the most appropriate translation, in order to induce \textit{chain-of-thought} reasoning effect~\cite{cot}.
        
        \begin{tcolorbox}[float, colback=gray!10, colframe=gray!80, title=Prompt 3: Translation Post-editing, label=box:prompt_post_edit, boxsep=1mm, left=2mm, right=2mm]
        \refstepcounter{mybox} \label{box:prompt_post_edit}
        \fontsize{10.5pt}{12pt}\selectfont
            You are a helpful assistant and an expert translator. You will be provided with a sentence in English and different possible translations in \highlight{\{Target Lang\}}.
            The English sentence can contain rich prosodic text with \highlight{\{Category-specific information\}}, that affects the meaning of the sentence.
            Your task is to select the most appropriate and prosody-aware translation.
            First provide a brief explanation of your reasoning and then the index of the selected translation.
    
            The sentence S to be translated is \highlight{\{sentence\}} and the candidate translations are:
            \highlight{$[\text{T}, \text{T}_\text{A}, \text{T}_\text{B}]$\}}
        
        \end{tcolorbox}
        
        After post-editing we remove all examples where the prosodic cases have identical translations, i.e.\ $(T_A{=}T_B)$. As an extra measure, we also remove examples where the word length-ratio of the non-prosodic translation $T$ and one of the prosodic translations $T_A, T_B$ is not within (0.75, 1.25)\footnote{We use character-based length-ratio for \texttt{Japanese}.}. This aims to remove translations that are overly explanatory, including new bits of information that can be due to the prosody, but are making the translation unnatural (see Table~\ref{tab:overly_expl_examples} in App.~\ref{app:overly_expl_examples} for examples.).

    \subsection{Controllable Speech Synthesis} \label{subsec:controllable_speech_synthesis}

        We use the OpenAI TTS which can synthesize very natural speech with high-quality audio, offering six different voice profiles. While there are no clear guidelines\footnote{\href{https://platform.openai.com/docs/guides/text-to-speech/overview}{platform.openai.com/docs/guides/text-to-speech}} on how to control prosody, we identified some effective prompting strategies to control the TTS output through trial-and-error~(Table \ref{tab:tts_prompting}).

        \begin{table}[h]
            \centering
            \resizebox{\columnwidth}{!}{
                \begin{tabular}{@{}cc@{}}
                \toprule
                \textbf{Effect}           & \textbf{TTS Prompting}                    \\ \midrule
                Strong Emphasis           & *WORD*                                    \\
                Normal Emphasis           & *word*                                    \\
                Slight Emphasis           & \_word\_                                  \\
                Pause                     & <pause>                                   \\
                Statement Intonation      & Prepend <statement>                       \\
                Question Intonation       & Prepend <question> \& Append ????      \\
                Emotional/Polite Tone & Prepend \& Append Emojis                  \\ \bottomrule
                \end{tabular}
            }
            \caption{OpenAI TTS prompting strategies.}
            \label{tab:tts_prompting}
        \end{table}

        To ensure that the generated audio follows the correct wording and exhibits the intended prosodic characteristics we use the following process: First, we generate six candidates (one per voice) for each prosody, discarding invalid candidates ($\text{WER}\!\neq\!0$) using an ASR model. Then we estimate prosody quality using category-specific tests in order to rank or filter examples. These tests employ techniques such as forced alignment~\cite{ctc_forced_alignment}, signal processing, punctuation probability, and speech emotion classification. They are explained in detail in Appendix~\ref{app:audio_filtering}.

\section{Contrastive Evaluation} \label{sec:contrastive_evaluation}

    General-purpose MT metrics like \textsc{BLEU} and \textsc{COMET} may be insensitive to subtle changes caused by prosody, and do not allow disentangling prosody awareness from overall translation quality. Thus, to assess how well an \stt{} model can handle prosody specifically, we develop a contrastive evaluation framework~\cite{contrastive_eval1}. Note that previous work on contrastive evaluation uses a single source and two or more targets \cite{contrastive_eval1,contrastive_eval2,prosody_st_korean} of which only one is correct. The model likelihood is then estimated for each target, and models are preferred that assign a better score to the correct example than to the foil(s). Here, we generalize this approach to leverage \benchmarkname{}'s \textit{double-contrastive} pairs, i.e.\ two sources and two targets (Fig.~\ref{tab:intro_examples}).

    Formally, each double-contrastive pair has two cases \(\{\textcolor{cyan}{X^a}, Z, \textcolor{cyan}{Y^a}\}\) and \(\{\textcolor{orange}{X^b}, Z, \textcolor{orange}{Y^b}\}\), where \(\textcolor{cyan}{X^a}, \textcolor{orange}{X^b}\) are the two different prosodic speech signals, \(Z\) is the source text (same for both cases), and \(\textcolor{cyan}{Y^a}, \textcolor{orange}{Y^b}\) are the different translated texts for each case. Thus, each example has two correct pairs \((\textcolor{cyan}{X^a}, \textcolor{cyan}{Y^a})\), \((\textcolor{orange}{X^b}, \textcolor{orange}{Y^b})\) and two incorrect ones \((\textcolor{cyan}{X^a}, \textcolor{orange}{Y^b})\), \((\textcolor{orange}{X^b}, \textcolor{cyan}{Y^a})\). We propose the following conditions to assess whether the \stt{} model can correctly solve the contrastive example, and to what degree:
    \begin{align*}
        \strong = \boldsymbol{1} \Big[ & f(\textcolor{cyan}{Y^a} \mid \textcolor{cyan}{X^a}; \theta) - f(\textcolor{orange}{Y^b} \mid \textcolor{cyan}{X^a}; \theta) > 0 \nonumber \\
        \text{and } & f(\textcolor{orange}{Y^b} \mid \textcolor{orange}{X^b}; \theta) - f(\textcolor{cyan}{Y^a} \mid \textcolor{orange}{X^b}; \theta) > 0 \Big] \\
        \weak = \boldsymbol{1} \Big[ & f(\textcolor{cyan}{Y^a} \mid \textcolor{cyan}{X^a}; \theta) - f(\textcolor{orange}{Y^b} \mid \textcolor{cyan}{X^a}; \theta) \nonumber \\
        + & f(\textcolor{orange}{Y^b} \mid \textcolor{orange}{X^b}; \theta) - f(\textcolor{cyan}{Y^a} \mid \textcolor{orange}{X^b}; \theta) > 0 \Big]
    \end{align*}

    Here, $\boldsymbol{1}[\cdot]$ is the indicator function, and $f(\cdot) > 0$ is a function that measures the agreement between audio input $X$ and target translation $Y$ under the \stt{} model with parameters $\theta$. $\strong$ is a \textit{global} condition, requiring the model to prefer both of the correct pairs versus the incorrect ones according to $f$.$\weak$ is a d\textit{irectional }condition~\cite{ribeiro-etal-2020-beyond} where we require a net positive directional movement for the two comparisons. We expect a model to have a strong internal representation of prosody if it can solve the global condition, and weak representation if it can only solve the directional one.\footnote{Note that $\strong$ is a sufficient condition for $\weak$.}

    We consider two different functions $f$ to measure the agreement of $X$ and $Y$.
    

    \subsection{Contrastive Likelihood} \label{subsec:contrastive_likelihood}

        Similar to prior work on contrastive evaluation~\cite{contrastive_eval1,contrastive_eval2,prosody_st_korean} we use the model likelihood to measure the level of agreement between input audio and target text. We obtain the model likelihood $\mathcal{L} \in \mathbb{R}^+$ for a reference $Y = (y_1, \dots, y_{|Y|})$, given a speech signal $X \in \mathbb{R}^k$ and an E2E \stt{} model with parameters $\theta_\text{E2E}$. It is defined as the product of the conditional probabilities, normalized by the length of the reference. Formally:
        \begin{align*}
            \mathcal{L}(Y \mid X; \theta_{\text{E2E}}) = \frac{1}{|Y|} \prod_{i=1}^{|Y|} p_{\theta_{\text{E2E}}} \big(y_i \mid X, y_{<i} \big)
        \end{align*}
        For a cascaded \stt{} model we approximate the true likelihood by considering the top-n \asr{} hypotheses $\mathcal{Z} = \{Z^{(1)}, \dots, Z^{(n)}\}$. Assuming the lengths of the $\mathcal{Z}$ are generally similar, we get:
        \begin{align}
            &\mathcal{L}(Y \mid X; \theta_{\text{casc}}) \approx \mathcal{L}(Y \mid \mathcal{Z}; \theta_{\text{MT}}) \mathcal{L}(\mathcal{Z} \mid X; \theta_{\text{ASR}}) \nonumber \\
            &\approx \frac{\sum_{j=1}^n \left[ \mathcal{L}(Y \mid Z^{(i)}; \theta_{\text{MT}}) \cdot \mathcal{L}(Z^{(i)} \mid X; \theta_{\text{ASR}}) \right]}{\sum_{j=1}^n \mathcal{L}(Z^{(i)} \mid X; \theta_{\text{ASR}})} \nonumber
        \end{align}
        Furthermore, to remove a potential bias of the model against rare translations, we normalize by the unconditioned decoder likelihood of the reference:\footnote{Estimated by using an empty audio for E2E case and empty source text in the \mt{} model for the cascade.}
        \begin{align}
            f_{\overline{\mathcal{L}}}(Y \mid X; \theta) &= \frac{\mathcal{L}(Y \mid X; \theta)}{\mathcal{L}(Y \mid \theta)} \label{eq:L_norm}
        \end{align}

    \subsection{Contrastive Translation Quality} \label{subsec:contrastive_xcomet}

        A common criticism of using model likelihoods is that they do not assess whether the correct output is actually generated in practice, due to teacher forcing. To address this, we propose another function that leverages translation quality estimation (QE) to compare unconstrained autoregressively generated model outputs. We obtain the hypothesis $\hat{Y}$ of input $X$ by generating with the \stt{} model $\mathcal{M}_\theta$, and use \textsc{xCOMET}~\cite{xcomet} to measure the quality of the translation. Thus:
        \begin{align}
            f_\mathcal{Q}(Y \mid X; \theta) = \mathcal{Q}\big(Y, \mathcal{M}_\theta(X)\big) = \mathcal{Q}(Y, \hat{Y}) \label{eq:qe}
        \end{align}
        The contrastive metrics using $f_\mathcal{Q}$ are expected to give us a better insight into how influential prosody is when translating with \stt{} models, as compared to using $f_\mathcal{L}$ (Eq.~\ref{eq:L_norm}), since they consider autoregressive generation and beam search.
    
\section{Experimental Setup} \label{sec:experimental_setup}

    \subsection{Data Generation} \label{subsec:exp_setup_data}

        For prosodic example generation with GPT-4 (\S\ref{subsec:prosodic_example_generation}) we used a temperature of 1, and 20 text domains per subcategory. The model was prompted to generate 10 examples\footnote{We generated 15/20 examples for intonation patterns/politeness, respectively.} for each pair of (subcategory, domain). The total number of subcategories is 27 (more details in App.~\ref{app:subcategories}), amounting to $5.5\text{k}$ examples of English sentences with pairs of prosodies and meanings created initially. Then we generated the candidates for the six voices with the TTS ($5.5\text{k}{\times}6{\times}2 = 66\text{k}$) and choose the $11\text{k}$ best candidates as described in \S\ref{subsec:controllable_speech_synthesis}. After quality assessment we end up with $2.8\text{k}$ examples with good prosody quality in the generated audio. Then we separately translated each one to the three target languages German (\texttt{De}), Spanish (\texttt{Es}), and Japanese (\texttt{Ja}). After post-editing and filtering we obtained $1.3\text{k}$--$1.4\text{k}$ full examples for each language pair (Table~\ref{tab:data_size}).

        \begin{table}[h]
            \centering
            \resizebox{0.99\columnwidth}{!}{
            \begin{tabular}{lccc}
            \toprule
            \textbf{Category} & \textbf{En-De} & \textbf{En-Es} & \textbf{En-Ja} \\
            \midrule
            Emotional prosody & 373 & 379 & 376 \\
            Sentence stress & 277 & 279 & 342 \\
            Prosodic breaks & 276 & 252 & 289 \\
            Politeness & 212 & 193 & 206 \\
            Intonation patterns & 173 & 173 & 173 \\
            \midrule
            \textbf{Total} & 1,311 & 1,294 & 1,386 \\
            \bottomrule
            \end{tabular}
            }
            \caption{Number of examples for each language pair in \benchmarkname{}. More details are in Appendix~\ref{app:data_statistics}.}
            \label{tab:data_size}
        \end{table}

    \subsection{Speech-to-text Translation Models} \label{subsec:exp_setup_models}

        We evaluated \stt{} models that fall under these three categories:
        \begin{itemize}\setlength\itemsep{-0.2em}
            \item E2E, where inference is done without an intermediate transcription step. The decoder of this model has full access to the prosody of the input.
            \item AED-based cascade, which is composed of an attentional encoder-decoder (AED)~\cite{transformer} \asr{} model and an \mt{} model. We expect the decoder of the MT model to have limited access to prosody, unless the ASR model is able to encode it in the transcription. This is possible mainly though punctuation, but also when the ASR model is acting more interpretative (i.e.\ generating synonyms that better fit the prosody rather than the spoken words).
            \item CTC-based cascade, which uses a CTC encoder~\cite{ctc} for the ASR part. The decoder of the MT model is expected to have almost no access to prosody since CTC model outputs are not punctuated and cannot be interpretative.
        \end{itemize}
        We are evaluating the following \stt{} models:
        \begin{itemize}\setlength\itemsep{-0.2em}
            \item \textsc{SeamlessM4T}~\citeSeamlessv{} is a multilingual and multimodal encoder-decoder. It is trained with multi-task learning on \asr{}, \mt{}, \stt{} and also on speech-to-speech translation (\sts{}), and can thus be used in either E2E or cascaded (AED) mode.
            \item XLS-R~\cite{xls-r} is a multilingual E2E model, of which the encoder is based on \textsc{wav2vec2.0} and its decoder on \textsc{mBART50}~\cite{mbart}.
            \item \textsc{ZeroSwot}~\cite{zeroswot} is a zero-shot E2E model that connects a \textsc{wav2vec 2.0} CTC encoder and NLLB~\citenllb{}.
            \item SALMONN~\cite{salmonn} is an audio LLM that connects \textsc{Whisper}~\cite{whisper} and BEATs~\cite{beats} to the Vicuna LLM~\cite{vicuna}, and can be used as an E2E \stt{} model.
            \item \textsc{Whisper} \& NLLB (AED-based cascade).
            \item CTC \& NLLB (CTC-based cascade) with \textsc{wav2vec 2.0} or \textsc{HuBERT}~\cite{hubert}.
        \end{itemize}

        We considered different versions of these 6 models, thus evaluating in total 31 \stt{} model variants of different sizes and capabilities (App.~\ref{app:st_models}).
        
    \subsection{Metrics} \label{sec:exp_setup_evaluation}
    
        We used beam search with beam size 5 to generate hypotheses. For estimating the conditional likelihood of the cascade (\S\ref{subsec:contrastive_likelihood}) we used the top-5 \asr{} hypotheses. For the contrastive translation quality (\S\ref{subsec:contrastive_xcomet}) we used \textsc{xCOMET-XL}\footnote{\href{https://huggingface.co/Unbabel/XCOMET-XL}{hf.co/Unbabel/XCOMET-XL}}~\cite{xcomet}, which is a state-of-the-art neural quality estimation metric based on XLM-R~\cite{xlm-r}. For all evaluated models we present their \textit{contrastive likelihood} and \textit{contrastive translation quality} scores, both \textit{global} and \textit{directional} versions, as a percentage of solved examples. We also evaluate them on standard QE using \textsc{xCOMET-XL}, by using the 2 correct pairs of each example ($2.6\text{k}$ samples). For statistical significance testing we used bootstrap resampling~\cite{bootstrap} with $10\text{k}$ resamples and a $95\%$ confidence interval.
    
\section{Experimental Results}  \label{sec:results}

    In Table~\ref{tab:main_results} we present the results of evaluating a selection of large and recent model versions all three language pairs. We find that most \stt{} models have at least some internal representation of prosody, enabling them to outperform the random baseline of 50$\%$ for the directional contrastive likelihood. On the other hand, when we consider autoregressive generation, we observe that the scores for the directional contrastive quality are relatively low\footnote{Assuming \textsc{xCOMET} is 0 for randomly generated text, the baseline scores are also 0.}, indicating that prosody is often not prominent enough in the internal representations of the models for it to be manifested in the generated translations. Furthermore, we find that the task of correctly solving both sub-cases of each example (global agreement) is very challenging for all models, with scores ranging around 10$\%$ for both contrastive metrics. We observe that even though the best performing model according to standard evaluation (\textsc{xCOMET}) is a cascade system, it falls behind the best E2E models when considering the contrastive evaluation on \benchmarkname{}. This finding illustrates why it is beneficial to separate prosody evaluation from general accuracy evaluation to study the phenomenon, which is further supported by our observation that the prosody and general accuracy metrics are only moderate correlated (see Fig.~\ref{fig:metric_correlation} in App.~\ref{app:results}).

    \begin{table*}[h]
    \centering
    \resizebox{0.8\textwidth}{!}{
    \begin{tabular}{@{}lccccc@{}}
    \toprule
    \multirow{2}{*}{\textbf{Model Name}}                               & \multicolumn{2}{c}{\textbf{Contrastive Likelihood}} & \multicolumn{2}{c}{\textbf{Contrastive Quality}} & \multirow{2}{*}{\textbf{\textsc{xCOMET}}} \\ \cmidrule(lr){2-5}
                                                                       & \textbf{Directional}           & \textbf{Global}           & \textbf{Directional}         & \textbf{Global}         &                                  \\ \midrule
    \multicolumn{6}{c}{\emph{English $\rightarrow$ German}}                                                                                                                                                     \\ \midrule
    \textsc{SeamlessM4T-v2-Large}                                        & 61.2                    & \textbf{13.5}             & 37.4                  & 14.5            & 0.988                            \\
    XLS-R 2B                                                           & 59.3                    & 4.6                       & 31.1                  & 7.3                     & 0.980                            \\
    \textsc{ZeroSwot-Large}                                              & 60.6                    & 9.7                       & 29.2                  & 8.7                     & 0.990                            \\
    SALMONN-13B                                                        & \textbf{62.8}           & 7.2                       & \textbf{43.2}         & \textbf{15.9}                     & 0.975                            \\ 
    \rowcolor{gray!10} \textsc{SeamlessM4T-v2-Large}                   & 60.2                    & 12.9                      & 31.1                  & 10.4                     & 0.991                            \\
    \rowcolor{gray!10} \textsc{Whisper-v3-Large} \& \textsc{NLLB-3.3B} & 60.7                    & 5.8                       & 23.1                  & 5.5                     & \textbf{0.992}                   \\
    \rowcolor{gray!10} \textsc{HuBERT-XL} \& \textsc{NLLB-3.3B}        & 39.4                    & 0.5                       & 20.5                  & 2.6                     & 0.979                            \\ \midrule
    \multicolumn{6}{c}{\emph{English $\rightarrow$ Spanish}}                                                                                                                                                      \\ \midrule
    \textsc{SeamlessM4T-v2-Large}                                          & \textbf{64.9}           & \textbf{13.4}             & 37.9                  & 11.0            & 0.982                            \\
    XLS-R 2B                                                           & 57.6                    & 5.6                       & 32.0                  & 8.4                     & 0.930                            \\
    \textsc{ZeroSwot-Large}                                             & 57.5                    & 9.2                       & 31.1                  & 5.6                     & 0.948                            \\
    SALMONN-13B                                                        & 61.3                    & 3.6                       & \textbf{39.6}         & \textbf{12.3}                      & 0.967                            \\ 
    \rowcolor{gray!10} \textsc{SeamlessM4T-v2-Large}                   & 61.3                    & 11.7                      & 29.5                  & 7.6                   & 0.984                            \\
    \rowcolor{gray!10} \textsc{Whisper-v3-Large} \& \textsc{NLLB-3.3B} & 63.2                    & 2.9                       & 25.4                  & 4.8                     & \textbf{0.987}                   \\
    \rowcolor{gray!10} \textsc{HuBERT-XL} \& \textsc{NLLB-3.3B}        & 41.8                    & 0.2                       & 20.8                  & 2.4                     & 0.968                            \\ \midrule
    \multicolumn{6}{c}{\emph{English $\rightarrow$ Japanese}}                                                                                                                                                       \\ \midrule
    \textsc{SeamlessM4T-v2-Large}                                           & 59.4                    & \textbf{12.4}             & 40.3                  & 13.8           & 0.956                            \\
    XLS-R 2B                                                           & 60.0                    & 4.6                       & 27.4                  & 7.0                     & 0.950                            \\
    \textsc{ZeroSwot-Large}                                            & 58.8                    & 7.9                       & 23.6                  & 7.9                     & \textbf{0.970}                   \\
    SALMONN-13B                                                        & \textbf{60.4}           & 10.8                      & \textbf{46.1}         & \textbf{16.1}                     & 0.859                            \\ 
    \rowcolor{gray!10} \textsc{SeamlessM4T-v2-Large}                   & 59.4                    & 9.1                       & 31.0                  & 8.7                     & 0.961                            \\
    \rowcolor{gray!10} \textsc{Whisper-v3-Large} \& \textsc{NLLB-3.3B} & 59.8                    & 4.9                       & 21.5                  & 5.3                     & 0.960                            \\
    \rowcolor{gray!10} \textsc{HuBERT-XL} \& \textsc{NLLB-3.3B}        & 40.4                    & 0.8                       & 15.7                  & 2.5                     & 0.922                            \\ \midrule
    \multicolumn{6}{c}{\emph{Average}}                                                                                                                                                                              \\ \midrule
    \textsc{SeamlessM4T-v2-Large}                                          & \textbf{61.8}           & \textbf{13.1}             & 38.5                  & 13.1          & 0.975                            \\
    XLS-R 2B                                                           & 59.0                    & 4.9                       & 30.2                  & 7.6                     & 0.953                            \\
    \textsc{ZeroSwot-Large}                                           & 59.0                    & 8.9                       & 28.0                  & 8.1                     & 0.969                            \\
    SALMONN-13B                                                        & 61.5                    & 7.2                       & \textbf{42.9}         & \textbf{14.8}                    & 0.933                            \\ 
    \rowcolor{gray!10} \textsc{SeamlessM4T-v2-Large}                   & 60.3                    & 11.2                      & 30.5                  & 8.9                     & 0.979                            \\
    \rowcolor{gray!10} \textsc{Whisper-v3-Large} \& \textsc{NLLB-3.3B} & 61.2                    & 4.5                       & 23.3                  & 5.2                     & \textbf{0.980}                   \\
    \rowcolor{gray!10} \textsc{HuBERT-XL} \& \textsc{NLLB-3.3B}        & 40.5                    & 0.5                       & 19.0                  & 2.5                     & 0.956                            \\ \bottomrule
    \end{tabular}
    }
    \caption{Contrastive Evaluation of \stt{} models on \benchmarkname{}. Grey background indicates a cascaded system.}
    \label{tab:main_results}
    \end{table*}

    \setlength{\parindent}{0pt}

\begin{figure*}[h]
        \centering
        \includegraphics[width=\textwidth]{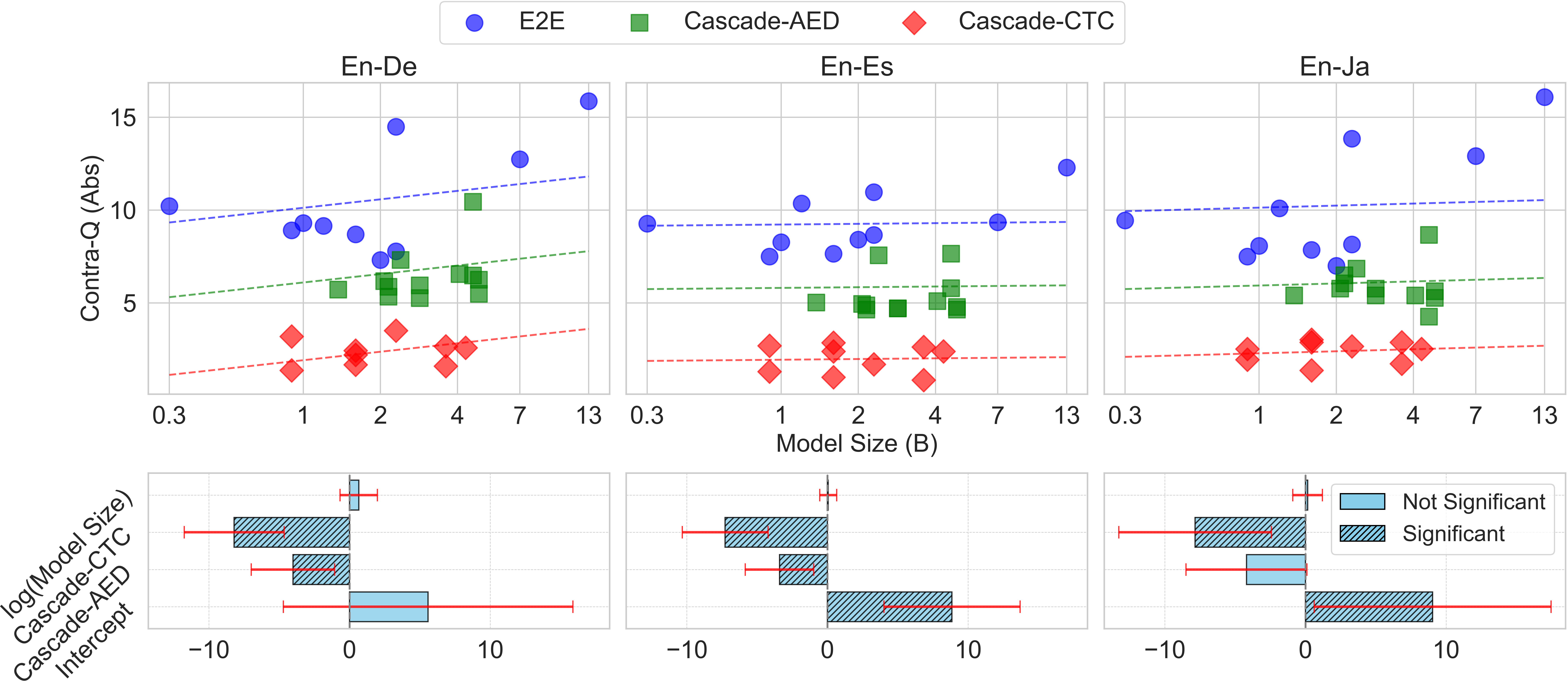}
        \caption{Regression Analysis of model types and model sizes per language pair.}
        \label{fig:regression_strong}
    \end{figure*}

    \textbf{Are model type and model size important for prosody-awareness?}
    We evaluate all 31 \stt{} models using \textit{global contrastive quality}, and run a regression analysis with the model type (E2E/AED-cascade/CTC-cascade) and model size as inputs. We use a mixed effects model~\cite{mixed_effects} to group together each model family, and thus account for random effects, such as the training data and hyperparameters. Specifically:
   \begin{align}
        y_{ij} \! = \! \beta_0 \! + \! \beta_1 \text{S}_{ij} \! + \! \beta_2\text{AED}_{ij} \! + \! \beta_3 \text{CTC}_{ij} \! + \! u_j \! + \! \epsilon_{ij}, \nonumber
    \end{align}
    where $y_{ij}$ is the score of $i$-th model variant of the $j$-th model family, $\beta_0$ is the intercept, $S$ is the log of the model size, AED and CTC are binary variables, $u_j$ is the random effect for $j$-th model family, and $\epsilon_{ij}$ is a residual error term.
    All scores are available in Table~\ref{tab:contra_q_scores_all} in App.~\ref{app:results}. In Figure~\ref{fig:regression_strong} we confirm with statistical significance that the E2E models outperform the cascades in all three language directions.\footnote{Note that results are borderline non-significant for \texttt{En}-\texttt{Ja} against the AED-cascade.} There is also a statistically significant negative impact on prosody-awareness when the cascade is based on a CTC ASR model that may be explained by the absence of punctuation in CTC transcripts, which if present can at least approximately signal some prosodic phenomena. Finally, although there is some evidence that larger models are more prosody-aware, results are not statistically significant. We speculate that larger models have more capacity to encode prosody in the weights, but since prosody is perhaps not sufficiently represented in the training data, this effect is limited.

\begin{figure*}[h]
        \centering
        \begin{minipage}[t]{0.9\textwidth}
            \centering
            \includegraphics[width=\textwidth]{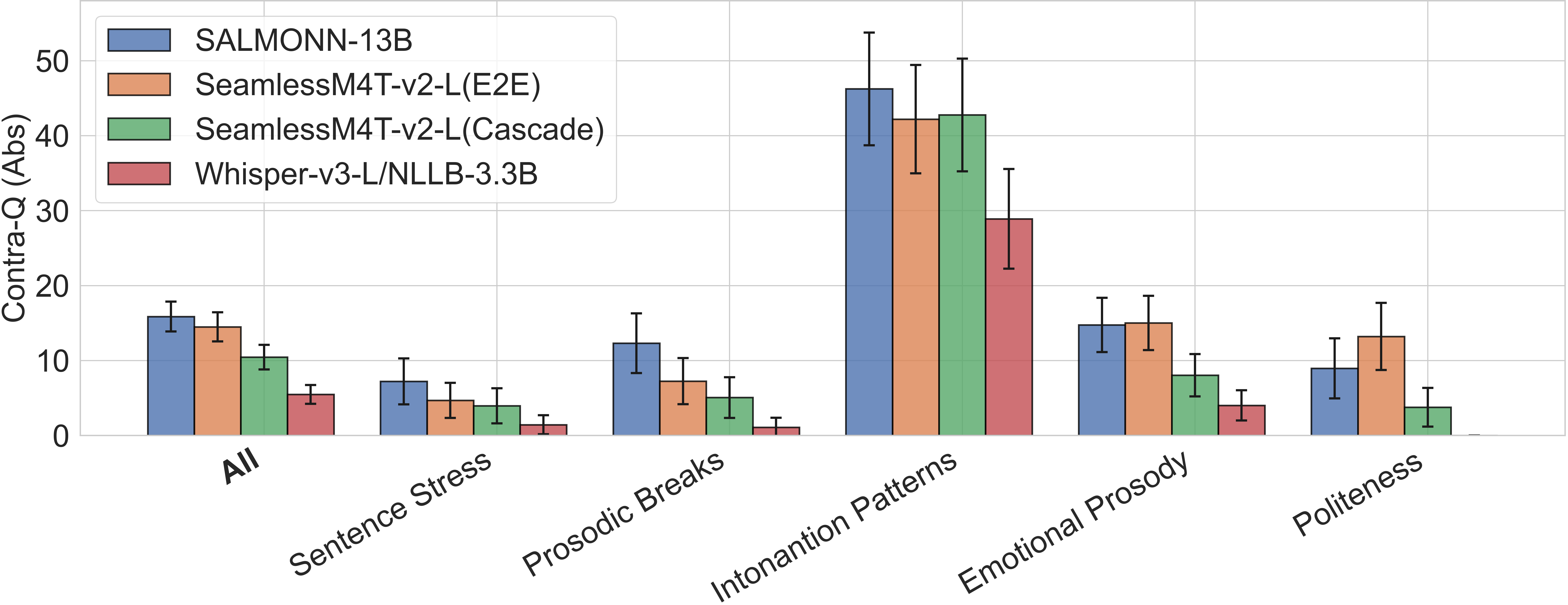}
        \end{minipage}
        \hfill
        \begin{minipage}[t]{\textwidth}
            \centering
            \includegraphics[width=0.8\textwidth]{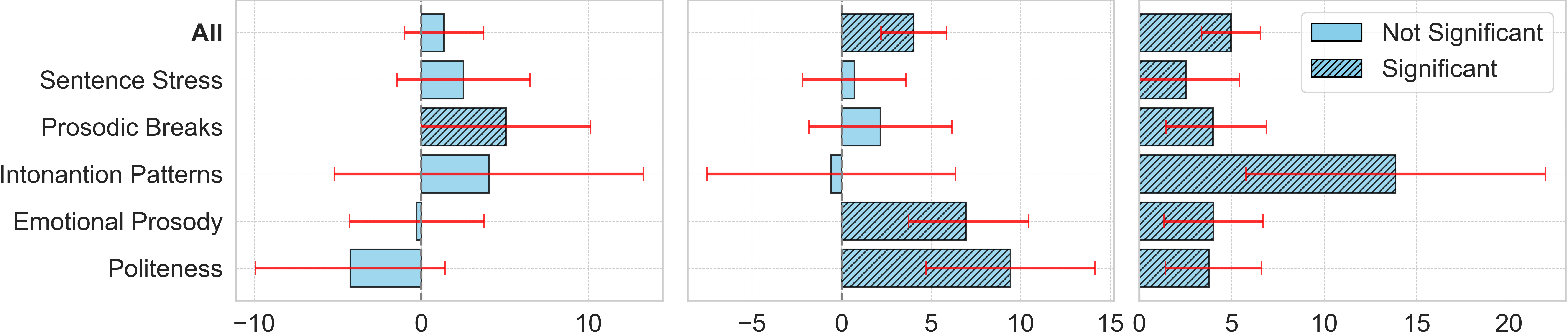}
        \end{minipage}
        \caption{Upper: Model performance per category (\texttt{En}-\texttt{De}). Lower Model performance comparisons (\texttt{En}-\texttt{De}), (a): SALMONN-13B vs.\ \textsc{SeamlessM4T-v2-Large}, (b) \textsc{SeamlessM4T-v2-Large}(E2E) vs.\ \textsc{SeamlessM4T-v2-Large}(cascade), (c) \textsc{SeamlessM4T-v2-Large}(cascade) vs.\ \textsc{Whisper-v3-Large}/NLLB-3.3B.}
        \label{fig:performance_comparisons_deu}
    \end{figure*}

    \textbf{How do results compare across categories and models?} In Figure~\ref{fig:performance_comparisons_deu} we present results across individual prosodic categories for four different English-German models, and perform pairwise model comparisons via bootstrap resampling\footnote{English-Spanish/Japanese are available at Figures~\ref{fig:performance_comparisons_spa}, \ref{fig:performance_comparisons_jpn} in App.~\ref{app:results}.}. The only category models are able to solve consistently is \textit{intonation patterns}, which can also be solved by cascaded models due to the presence of punctuation in the transcription. The comparably lower scores in the other four categories further demonstrate the inability of current state-of-the-art models to use prosody, with \textit{sentence stress} being the most challenging. Through the pairwise comparisons, we find that an LLM-based model (SALMONN) is not statistically different from a more standard \stt{} model, like \textsc{SeamlessM4T}. Next, comparing the \textsc{SeamlessM4T} model in both E2E and cascade allows us to control for parameters such as training data and architecture, in order to observe the effect of model type, giving more clarity of our results on the theoretical advantage of E2E models. Finally, we observe a clear performance gain by using the \textsc{SeamlessM4T} cascade over the \textsc{Whisper} \& NLLB one. We hypothesize this advantage is due to the multitasking nature of \textsc{SeamlessM4T}, which makes its ASR mode more interpretative than standard ASR models. This allows the ASR part of the cascade to escape the word-by-word paradigm, and use more fitting words in the transcription (such as synonyms) that fit better the prosody of the audio. Supporting this hypothesis. we observe a worse WER score for \textsc{SeamlessM4T} ($11\%$) compared to \textsc{Whisper} ($4\%$).

\begin{figure}[h]
        \centering
        \includegraphics[width=0.99\columnwidth]{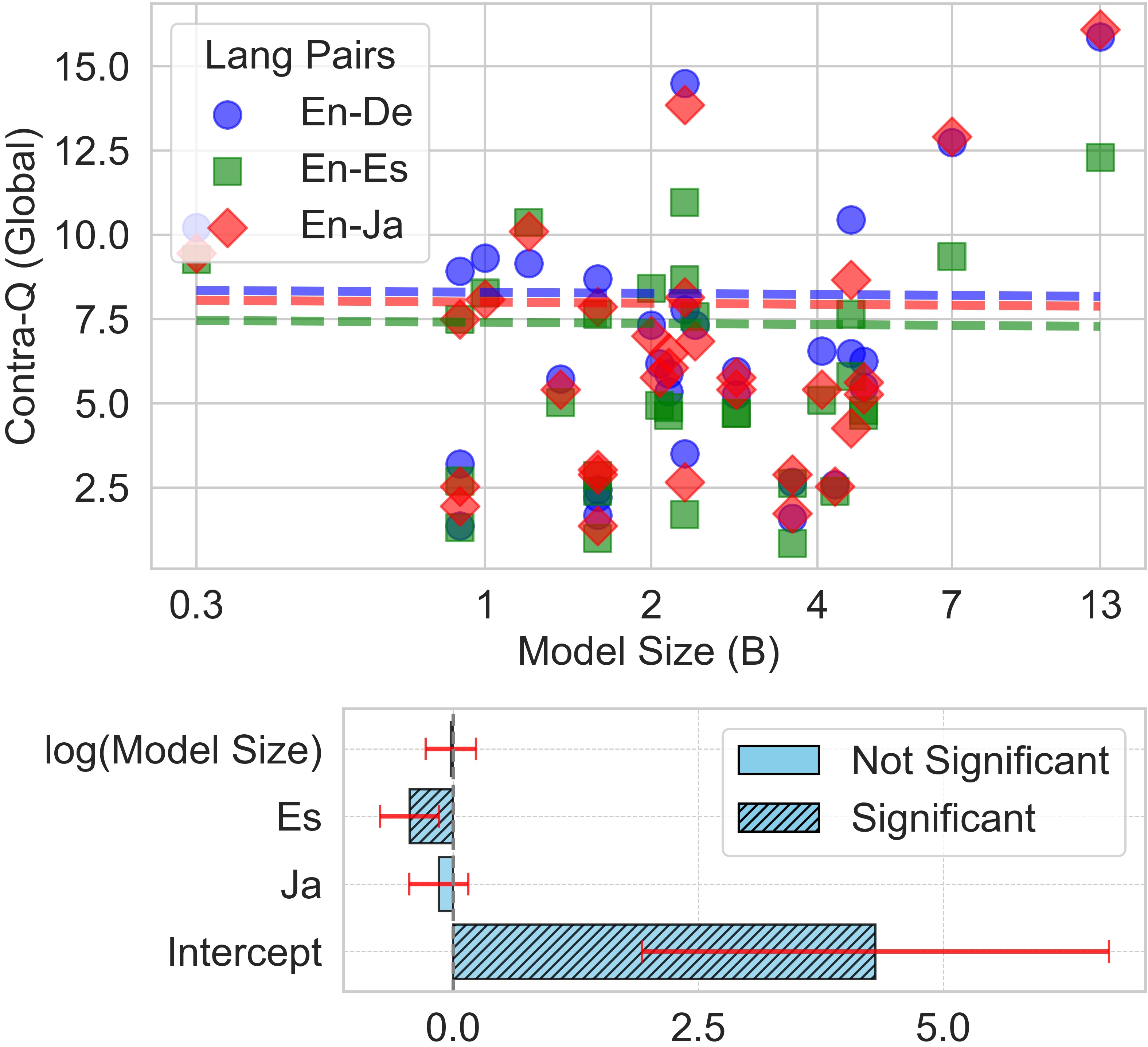}
        \caption{Regression Analysis of language pairs.}
        \label{fig:regression_lang_strong}
        \vspace{-0.1cm}
    \end{figure}

    \textbf{Is the level of prosody-awareness language-dependent?} In Figure~\ref{fig:regression_lang_strong} we carry out a similar regression analysis as in Figure~\ref{fig:regression_strong}, but with the language pair as an independent categorical variable. Interestingly, we observe that there are differences between the three language pairs, and also significant for Spanish vs.\ German, which indicates that prosody-awareness in \stt{} could be language-dependent. We hypothesize that the expressivity of the target language might be a relevant factor, since more expressive languages might be able to easier encode the prosody of the source speech into text.
    
    \setlength{\parindent}{12pt}

\section{Related Work} \label{sec: related_work}

    Prosody has traditionally been an important topic for TTS research~\cite{tts_prosody0}, either for transferring~\cite{tts_prosody1} or encoding it~\cite{tts_prosody2} in the synthesized speech. Furthermore, \citet{adept} created a dataset for evaluating prosody transfer in TTS models, which contains several categories, similar to our study here. Naturally prosody has also been the focus of \sts{} systems, in order to translate in a more expressive way~\cite{s2s_prosody1,s2s_prosody2,seamless2}. The topic has received less attention in the context of \stt{}. \citet{meld_st} present a dataset for emotional prosody based on speech and translations from TV series, and show that finetuning with emotion labels, can improve translation quality. \citet{prosody_st_korean} studied the prosody-awareness of \textsc{Whisper} in E2E and cascade mode, in translating Korean \textit{wh-phrases} using contrastive likelihood, and find evidence of the E2E model outperforming the cascade. Here we contribute a broader study of prosody in \stt{}, by proposing a double-contrastive benchmark that covers several prosodic categories, the use of more generative-like contrastive evaluation, and evaluating a plethora of \stt{} models. Finally, \citet{prosaudit} present a benchmark for evaluating prosody-awareness in self-supervised acoustic representations. Similarly to our study they present evidence of prosody awareness in the representations. Contrary to our results, they conclude that size has a positive effect on prosody awareness.

\section{Conclusions} \label{sec:conclusions}

    We presented \benchmarkname{}, a benchmark based on double-contrastive examples for evaluating prosody-awareness in \stt{} models, covering several categories and languages. In addition to standard contrastive evaluation based on model likelihoods, we proposed a generative contrastive metric based on quality estimation. We evaluated a plethora of models, and found that they exhibit some signs of prosody-awareness, but the effect is often not strong enough to influence the translations. We also confirmed the previously hypothesized inherent advantage of E2E models compared to cascaded models. We hope that our benchmark and findings will motivate more research into prosody-aware \stt{} in the future, enabling us to better understand it and improve it.


\section*{Limitations}

    For creating \benchmarkname{} we relied on an almost entirely automated data generation process. This allowed us to create a comprehensive dataset covering several prosodic phenomena and three language pairs, in a fast and cost-effective way.
    It would also enable expanding the coverage of languages and prosodic phenomena relatively easy in the future. Nevertheless, despite our best efforts regarding filtering and quality assessment (\S\ref{sec:data_gen} and App.~\ref{app:audio_filtering}), the data is not perfect and includes a certain amount of noise. We observed the following sources of noise in order of decreasing importance: (1) prosody not prominent in the generated speech; (2) translations overly explanatory or not encoding prosody; (3) semantic interpretations of the two cases rather similar. We do not expect these issues to be so frequent as to alter the findings of this work in a systematic way, but additional human annotation or verification would be a valuable step for future work. Furthermore, as the landscape of available generative models, in particular controllable TTS, is changing quickly, the quality of results using our data generation process would expectantly become less of a concern in future iterations. 

    Our study follows a contrastive evaluation methodology in order to isolate prosody-related behavior. As a consequence, our study does not allow drawing conclusions on how much prosody matters in real life data, and in what domains it is especially important. In addition, we hypothesize that some prosodic phenomena could be correctly translated by having access to the broader context of the conversation (context-aware \stt{}), which we leave for future research.

\newpage

\bibliography{acl_latex}

\appendix

\section{Prosodic Subcategories}
\label{app:subcategories}

    Here we expand the categorization of \S\ref{subsec:categories}, and discuss the identified subcategories for sentence stress and prosodic breaks, which are 4 and 6 respectively. Intonation patterns and Politeness do not have subcategories. For emotional prosody we have 15 emotion pairs\footnote{Removed \emph{fearful} emotion due to issues with the TTS.}, thus having 15 subcategories. Examples are available at Tables~\ref{tab:examples_stress} and \ref{tab:examples_pause}.

    \subsection{Sentence Stress Subcategories}
    \label{app:subcategories_sentence_stress}

        \setlength{\parindent}{0pt}
        
        (1.1) \emph{Contrastive Stress}, which highlights differences or corrects previous statements, emphasizing contrasts between elements~\cite{Bolinger1961}.
        
        (1.2) \emph{New vs.\ Given Information}, which differentiates between new and given information, emphasizing what is considered new~\cite{Halliday1967}.
        
        (1.3) \emph{Relational vs.\ Descriptive Adjectives}, where stressing the adjective or the noun can differentiate between the relational and descriptive uses of attributive adjectives~\cite{Sproat1992}.
    
        (1.4) \emph{Focus-Sensitive Operators}, where stress indicates the focus of adverbs of quantification (\emph{only}, \emph{just}, etc), shifting the meaning of the sentence accordingly~\cite{Halliday1967,Jackendoff1972}.

        \setlength{\parindent}{12pt}

    \subsection{Prosodic Break Subcategories}
    \label{app:subcategories_prosodic_breaks}

        \setlength{\parindent}{0pt}
    
        (2.1) \emph{Direct vs.\ Indirect Statements}, where a prosodic break can indicate whether a phrase is a direct or an indirect quote~\cite{direct_vs_indirect1,direct_vs_indirect2}.
        
        (2.2) \emph{Restrictive vs.\ Non-Restrictive Clauses}, which involves the use of prosodic breaks to differentiate between essential and non-essential information, impacting the specificity of the noun being described~\cite{Nespor1986}.
        
        (2.3) \emph{VP vs.\ NP Attachment}, where a trailing phrase can be attached either to the verb-phrase or the noun-phrase, depending on the existence of a prominent prosodic break~\cite{Pynte1996}.
    
        (2.4) \emph{Particle vs.\ Preposition}, where a prosodic break can disambiguate between the literal and idiomatic meaning of phrasal verbs, by grouping the preposition with or without it~\cite{Price1991}.
    
        (2.5) \emph{Broad vs.\ Narrow Scope}, where the existence of a prosodic break can signal that a modifier (adjective) has narrow scope, and refers only to one of two nouns that follow it~\cite{Hirschberg1992}.
    
        (2.6) \emph{Complementizer vs.\ Parenthetical}, where the location of a prosodic break indicates whether an intermediate phrase acts as a complementizer or simply parenthetical to the main one~\cite{Dehe2014}.

        \setlength{\parindent}{12pt}

\section{Examples for In-context Learning} \label{app:examples}

    In Tables \ref{tab:examples_stress}, \ref{tab:examples_pause} and \ref{tab:examples_rest} we present some of the examples used for in-context learning when generating new examples with GPT-4 (\S\ref{subsec:prosodic_example_generation}).

    \begin{table*}[]
        \centering
        \resizebox{\textwidth}{!}{
        \begin{tabular}{ll}
        \toprule
        \multicolumn{2}{c}{\textbf{1.1 Contrastive Stress}}                                                              \\
        Sentence    & She didn't give the book to John.                                                                \\
        Prosody$_\text{A}$ & She didn't give the *BOOK* to John.                                                              \\
        Meaning$_\text{A}$ & Something else was given to John.                                                                \\
        Prosody$_\text{B}$ & She didn't give the book to *JOHN*.                                                              \\
        Meaning$_\text{B}$ & The book was given to someone else.                                                              \\ \hline
        \multicolumn{2}{c}{\textbf{1.2 New vs.\ Given Information}}                                                          \\
        Sentence    & The committee decided to postpone the meeting.                                                   \\
        Prosody$_\text{A}$ & The *COMMITTEE* decided to postpone the meeting.                                                 \\
        Meaning$_\text{A}$ & Given: Someone decided to postpone the meeting; New: It was the committee who decided.           \\
        Prosody$_\text{B}$ & The committee decided to *POSTPONE* the meeting.                                                 \\
        Meaning$_\text{B}$ & Given: The committee decided something; New: The decision was to postpone it.                    \\ \hline
        \multicolumn{2}{c}{\textbf{1.3 Relational vs.\ Descriptive Adjectives}}                                            \\
        Sentence    & They are German teachers.                                                                        \\
        Prosody$_\text{A}$ & They are *GERMAN* teachers.                                                                      \\
        Meaning$_\text{A}$ & Teachers who teach the German language. (Relational)                                             \\
        Prosody$_\text{B}$ & They are German *TEACHERS*.                                                                      \\
        Meaning$_\text{B}$ & Teachers who are German. (Descriptive)                                                           \\ \hline
        \multicolumn{2}{c}{\textbf{1.4 Focus-Sensitive Operators}}                                                         \\
        Sentence    & I only introduced John to Maria at yesterday's party.                                            \\
        Prosody$_\text{A}$ & I only introduced *JOHN* to Maria at yesterday's party.                                          \\
        Meaning$_\text{A}$ & John was the only person I introduced to Maria.                                                  \\
        Prosody$_\text{B}$ & I only introduced John to *MARIA* at yesterday's party.                                          \\
        Meaning$_\text{B}$ & Maria was the only person I introduced John to.                                                  \\ \hline
        \end{tabular}
        }
        \caption{Examples in the category \emph{Sentence Stress} that were used for in-context learning.}
        \label{tab:examples_stress}
    \end{table*}
    
    \begin{table*}[h!]
        \centering
        \resizebox{\textwidth}{!}{
        \begin{tabular}{ll}
        \toprule
        \multicolumn{2}{c}{\textbf{2.1 Direct vs.\ Indirect Statements}}                                                             \\
        Sentence    & Alex announced Jamie will meet the manager.                                                               \\
        Prosody$_\text{A}$ & Alex *ANNOUNCED* $|$ Jamie will meet the manager.                               \\
        Meaning$_\text{A}$ & (Direct Statement)                                                                          \\
        Prosody$_\text{B}$ & Alex announced Jamie will meet the manager.                                                               \\
        Meaning$_\text{B}$ & (Indirect Statement)                                                                             \\ \hline
        \multicolumn{2}{c}{\textbf{2.2 Restrictive vs.\ Non-Restrictive Phrases}}                                                    \\
        Sentence    & The students who were talking were sent out.                                                             \\
        Prosody$_\text{A}$ & The students who were *TALKING* $|$ were sent out.                              \\
        Meaning$_\text{A}$ & Only the students who were talking were actually sent out. (Restrictive)                                 \\ 
        Prosody$_\text{B}$ & The *STUDENTS* $|$ who were talking $|$ were sent out. \\
        Meaning$_\text{B}$ & All students were sent out, and the fact they were talking is additional information. (Non-restrictive)  \\ \hline
        \multicolumn{2}{c}{\textbf{2.3 Verb-phrase vs.\ Noun-phrase Attachment}}                                                 \\
        Sentence    & Paula phoned her friend from Alabama.                                                                    \\
        Prosody$_\text{A}$ & Paula phoned her friend $|$ from *ALABAMA*.                                     \\
        Meaning$_\text{A}$ & Paula called her friend while she was in Alabama. (VP Attachment)                  \\
        Prosody$_\text{B}$ & Paula phoned $|$ her *FRIEND* from Alabama.                                                                  \\
        Meaning$_\text{B}$ & Paula phoned her friend who is from Alabama. (NP Attachment)                  \\ \hline
        \multicolumn{2}{c}{\textbf{2.4 Phrasal Verbs}}                                                                     \\
        Sentence    & John laughed at the party.                                                                       \\
        Prosody$_\text{A}$ & John *LAUGHED* $|$ at the party.                                                                     \\
        Meaning$_\text{A}$ & John laughed while he was at the party. (Literal)                                                \\
        Prosody$_\text{B}$ & John *LAUGHED AT* $|$ the party.                                                                     \\
        Meaning$_\text{B}$ & John made fun of the party. (Idiomatic)                                                          \\  \hline
        \multicolumn{2}{c}{\textbf{2.5 Complementizer vs.\ Parenthetical}}                                      \\
        Sentence    & We only suspected they all knew that a burglary had been committed.        \\
        Prosody$_\text{A}$ & We only *SUSPECTED* $|$ they all knew that a burglary had been committed.     \\
        Meaning$_\text{A}$ & The suspicion was that they all knew about the burglary. (Complementizer)    \\
        Prosody$_\text{B}$ & We only suspected $|$ they all *KNEW* $|$ that a burglary had been committed.           \\
        Meaning$_\text{B}$ & They all knew that we only suspected that a burglary had been committed. (Parenthetical)      \\  \hline
        \multicolumn{2}{c}{\textbf{2.6 Modifier Scope}}                                      \\
        Sentence    & This collar is dangerous to younger dogs and cats.        \\
        Prosody$_\text{A}$ & This collar is dangerous to *YOUNGER* dogs and cats.     \\
        Meaning$_\text{A}$ & Younger refers to both dogs and cats. (Broad Scope)    \\
        Prosody$_\text{B}$ & This collar is dangerous to *YOUNGER* dogs $|$ and *CATS*.   \\
        Meaning$_\text{B}$ & Younger refers only to dogs. (Narrow Scope)     \\  \bottomrule
        \end{tabular}
        }
        \caption{Examples in the category \emph{Prosodic Breaks} that were used for in-context learning.}
        \label{tab:examples_pause}
    \end{table*}
    
    \begin{table*}[h!]
        \centering
        \resizebox{\textwidth}{!}{
        \begin{tabular}{ll}
        \toprule
        \multicolumn{2}{c}{\textbf{3. Intonation Patterns}}                                                              \\
        Sentence    & You can solve this problem                                                                \\
        Prosody$_\text{A}$ & You *CAN* solve this problem.                                                              \\
        Meaning$_\text{A}$ & Encouraging or asserting the person's ability to solve this problem.                                                                \\
        Prosody$_\text{B}$ & You \_can\_ solve this problem?                                                              \\
        Meaning$_\text{B}$ & Questioning the person's ability to solve this problem.                                                              \\ \hline
        \multicolumn{2}{c}{\textbf{4. Emotional Prosody (Happy/Sad)}}                                                          \\
        Sentence    & The surgery went as expected.                                                   \\
        Prosody$_\text{A}$ & $<$happy$>$ The surgery went *AS EXPECTED*!                                                 \\
        Meaning$_\text{A}$ & The surgery's successful outcome aligns with hopes and predictions, leading to joy and relief.           \\
        Prosody$_\text{B}$ & $<$sad$>$ The surgery went \_as expected\_ ...                                                 \\
        Meaning$_\text{B}$ & The expected outcome was not favorable, leading to a somber tone.                    \\ \hline
        \multicolumn{2}{c}{\textbf{4. Emotion Prosody (Fearful/Angry)}}                                            \\
        Sentence    & Can we talk about this later?                                                                        \\
        Prosody$_\text{A}$ & $<$fearful$>$ Can we... talk about this... later?                                                                      \\
        Meaning$_\text{A}$ & Indicates hesitation or fear about the topic, or the situation in general.                                             \\
        Prosody$_\text{B}$ & $<$angry$>$ Can we *TALK* about this later!?                                                                      \\
        Meaning$_\text{B}$ & Implies urgency or frustration, and a demand for immediate attention.                                                           \\ \hline
        \multicolumn{2}{c}{\textbf{5. Politeness}}                                                        \\
        Sentence    & Can you move your car?                                            \\
        Prosody$_\text{A}$ & $<$polite$>$ Can you \_move\_ your car?                                         \\
        Meaning$_\text{A}$ & A polite request to move the car.                                                  \\
        Prosody$_\text{B}$ & $<$impolite$>$ Can you *MOVE* your *CAR*?!                                          \\
        Meaning$_\text{B}$ & A rude demand to move the car, with an aggressive tone.                                                  \\ \bottomrule
        \end{tabular}
        }
        \caption{Examples in the categories \emph{Intonation Patterns}, \emph{Emotional Prosody}, \emph{Politeness}, and that were used for in-context learning.}
        \label{tab:examples_rest}
    \end{table*}
    
\section{Quality Assessment for TTS candidates} \label{app:audio_filtering}

    Here we present the objectives we defined for assessing the quality of the generated speech candidates for each contrastive example. The objective is applied only to candidates that had $\text{WER} \! = \! 0$ using \textsc{Whisper}. If all candidates are invalid for a prosodic case, the whole example is removed. We also defined some threshold levels for the objectives after trial-and-error, in order to remove examples where the best candidate was below it.

    \setlength{\parindent}{0pt}

    \textbf{Sentence Stress.} We use forced-alignment with \textsc{wav2vec 2.0}~\cite{wav2vec2} to obtain the segment for each word in the signal, and extract their loudness, pitch and duration features. Then we define the stress level $\textit{stress}$ for a word $w$ as the weighted sum of these three features. Finally we select the best candidate according to a simple objective $\textit{obj}_{stress}$ that has three goals: (1) maximize the stress of the target word ($\textit{stress}_\textit{tgt}$), (2) minimize the stress of the target word of the contrastive case ($\textit{stress}_\textit{foil}$), and (3) minimize the average stress of the rest.
    \begin{align}
        \textit{stress}_w &= \lambda_1 \textit{loud}_w + \lambda_2 \textit{pitch}_w + \lambda_3 \textit{dur}_w \nonumber \\
        \textit{obj}_\textit{stress} &= 2 \cdot \textit{stress}_\textit{tgt}- \textit{stress}_\textit{foil} \nonumber \\ &- \frac{1}{n-1}\sum_{w \neq \textit{tgt}} \textit{stress}_w, \nonumber
    \end{align}
    where we used $\lambda_1 \! = \! 0.5$, $\lambda_2 \! = \! 0.3$, and  $\lambda_3 \! = \! 0.2$. Note that in the sentence stress examples, there is always exactly 1 target word in each contrastive prosodic case.

    \textbf{Prosodic Breaks.} Likewise, after forced-alignment, we measure the duration \textit{dur} of each gap $l$ between the words in the utterance, and define a similar objective $\textit{obj}_\textit{break}$ as:
    \begin{align}
        \textit{obj}_\textit{break} &= 2 \frac{1}{|\textit{tgt}|} \sum_{l \in \textit{tgt}} \textit{dur}_l - \frac{1}{|\textit{foil}|} \sum_{l \in \textit{foil}} \textit{dur}_l \nonumber \\
        &- \frac{1}{n-|\textit{tgt}|}\sum_{l \notin \textit{tgt}} \textit{dur}_l \nonumber
    \end{align}
    In this category, there can be 0 to 2 breaks in each prosodic case, which could be shared between the two prosodic cases. In the objective we consider only the ones that are not common in the two cases.

    \textbf{Intonation Patterns.} We use teacher-forcing with \textsc{Whisper} to extract the punctuation probabilities given the transcription text without the ending punctuation. The probability of the sentence to be a statement is the sum of the probabilities of the tokens ``.'' and ``!'', while the probability of a question is the probability of the token ``?''. Thus the objective $\textit{obj}_{inton}$ for a statement is defined as:
    \begin{align}
        \textit{obj}_{\textit{inton}} &= p(. \mid X, Z_{<n}) + p(! \mid X, Z_{<n}) \nonumber \\
        &- p(? \mid X, Z_{<n}), \nonumber
    \end{align}
    where $X$ is the speech signal and $Z_{<n}$ are the tokens of the transcription, excluding the final one, which corresponds in all cases of this category. to the punctuation. The negative objective $-\textit{obj}_{\textit{inton}}$ is used for a case that is a question.

    \textbf{Emotional Prosody.} We employ an emotion classifier\footnote{\href{https://huggingface.co/ehcalabres/wav2vec2-lg-XLS-R-en-speech-emotion-recognition}{hf.co/ehcalabres/wav2vec2-lg-XLS-R-en-speech-emotion-recognition}} which is a based on a finetuned \textsc{wav2vec 2.0} on the RAVDESS dataset~\cite{ravdess}, and define the objective as:
    \begin{align}
        \textit{obj}_{\textit{emo}} &= p(e_\textit{tgt} \mid X) - p(e_\textit{foil} \mid X), \nonumber
    \end{align}
    where $\theta$ are the parameters of the classifier, $e_\textit{tgt}$ is the target emotion label and $e_\textit{foil}$ is the emotion label of the other prosodic case.

    \textbf{Pragmatic Prosody.} To the best of our knowledge there is no open-sourced audio classifier to detect politeness levels, thus we re-purpose the emotion classifier and define the probabilities of politeness and impoliteness as a weighted sum of the 8 available emotion classes.
    \begin{align}
        p(\textit{polite}) &= \frac{\sum_{e} w_e p(e \mid X)}{\sum_{e} w_e}, \nonumber
    \end{align}
    and similarly for impolite. We used the weighted scheme displayed in Table~\ref{tab:polite_weights}, which was obtained by prompting GPT-4.

    \begin{table}[h!]
    \centering
    \resizebox{0.8\columnwidth}{!}{
    \begin{tabular}{ccc}
    \toprule
    \textbf{Emotion} & \textbf{Politeness} & \textbf{Impoliteness} \\ \midrule
    Happy     & 0.3  & -0.1 \\
    Calm      & 0.3  & -0.2 \\
    Neutral   & 0.2  & 0.1  \\
    Surprised & 0.1  & 0.1  \\
    Sad       & 0.0  & 0.2  \\
    Disgust   & -0.1 & 0.3  \\
    Angry     & -0.2 & 0.4  \\
    Fearful   & -0.1 & 0.0  \\ \bottomrule
    \end{tabular}
    }
    \caption{Weighting scheme for Politeness and Impoliteness labels based on the emotion classifier.}
    \label{tab:polite_weights}
    \end{table}

    \setlength{\parindent}{12pt}

\section{Data}

\subsection{Data Statistics}
\label{app:data_statistics}

    In Table \ref{tab:data_size_extended} we provide the analytic data statistics for each category/subcategory, throughout the generation process stages. The poor quality of the cTTS, where prosody was not always encoded in the speech, led us to remove a large percentage of the examples before translating them. Also many examples where removed because the oracle translations for both cases were the same.

    \begin{table*}[h!]
    \centering
    \resizebox{0.8\textwidth}{!}{
    \begin{tabular}{@{}lcccccc@{}}
    \toprule\toprule
     & \multirow{2}{*}{\textbf{Initial}} & \multirow{2}{*}{\textbf{Generated}} & \multirow{2}{*}{\textbf{Synthesised}} & \multicolumn{3}{c}{\textbf{Translated}}       \\ \cmidrule(l){5-7} 
       \textbf{Category / Subcategory}                                              &                                   &                                     &                                       & \textbf{De}   & \textbf{Es}   & \textbf{Ja}   \\ \midrule \midrule
    Contrastive Stress (General)                     & 200                               & 199                                 & 183                                   & 87            & 76            & 97            \\
    Relational/Descriptive Adjectives                & 200                               & 199                                 & 147                                   & 42            & 33            & 51            \\
    Contrastive Stress (Noun-Phrase)                 & 200                               & 199                                 & 124                                   & 37            & 36            & 39            \\
    New/Given Information                            & 200                               & 197                                 & 146                                   & 51            & 65            & 91            \\
    Focus-sensitive Operators                        & 200                               & 181                                 & 118                                   & 60            & 42            & 64            \\ \cmidrule(l){2-7} 
    \textbf{Sentence Stress}                         & {\ul 1000}                        & {\ul 975}                           & {\ul 718}                             & {\ul 277}     & {\ul 252}     & {\ul 342}     \\ \midrule\midrule
    Complementizer/Parenthetical                     & 200                               & 200                                 & 171                                   & 59            & 46            & 73            \\
    VP/NP Attachment                                 & 200                               & 200                                 & 66                                    & 23            & 18            & 20            \\
    Modifier Scope                                   & 200                               & 200                                 & 200                                   & 83            & 107           & 81            \\
    Restrictive/Nonrestrictive                       & 200                               & 199                                 & 177                                   & 65            & 82            & 40            \\
    Direct/Indirect                                  & 200                               & 198                                 & 154                                   & 41            & 25            & 70            \\
    Phrasal Verbs                                    & 42                                & 42                                  & 17                                    & 5             & 1             & 5             \\ \cmidrule(l){2-7} 
    \textbf{Prosodic Breaks}                         & {\ul 1042}                        & {\ul 1039}                          & {\ul 785}                             & {\ul 276}     & {\ul 279}     & {\ul 289}     \\ \midrule\midrule
    \textbf{Intonation Patterns}                     & {\ul 300}                         & {\ul 263}                           & {\ul 174}                             & {\ul 173}     & {\ul 173}     & {\ul 173}     \\ \midrule\midrule
    Sad-Happy                                        & 200                               & 200                                 & 1                                     & 1             & 1             & 1             \\
    Neutral-Angry                                    & 200                               & 199                                 & 185                                   & 123           & 111           & 119           \\
    Neutral-Happy                                    & 200                               & 198                                 & 161                                   & 81            & 97            & 81            \\
    Disgust-Angry                                    & 200                               & 198                                 & 18                                    & 4             & 5             & 3             \\
    Disgust-Sad                                      & 200                               & 198                                 & -                                     & -             & -             & -             \\
    Neutral-Surprised                                & 200                               & 198                                 & 43                                    & 33            & 35            & 30            \\
    Disgust-Neutral                                  & 200                               & 197                                 & 7                                     & 2             & 5             & 5             \\
    Happy-Angry                                      & 200                               & 197                                 & 138                                   & 50            & 65            & 72            \\
    Sad-Surprised                                    & 200                               & 197                                 & 3                                     & 2             & 2             & 2             \\
    Sad-Neutral                                      & 200                               & 196                                 & 4                                     & 3             & 2             & 2             \\
    Sad-Angry                                        & 200                               & 196                                 & 5                                     & 1             & 4             & 4             \\
    Disgust-Surprised                                & 200                               & 196                                 & 4                                     & 2             & 2             & 1             \\
    Disgust-Happy                                    & 200                               & 195                                 & 10                                    & 5             & 7             & 6             \\
    Happy-Surprised                                  & 200                               & 195                                 & 52                                    & 34            & 27            & 21            \\
    Angry-Surprised                                  & 200                               & 193                                 & 68                                    & 32            & 34            & 30            \\ \cmidrule(l){2-7} 
    \textbf{Emotional Prosody}                       & {\ul 3000}                        & {\ul 2953}                          & {\ul 699}                             & {\ul 433}     & {\ul 418}     & {\ul 377}     \\ \midrule\midrule
    \textbf{Politeness}                              & {\ul 400}                         & {\ul 375}                           & {\ul 387}                             & {\ul 212}     & {\ul 193}     & {\ul 206}     \\ \midrule\midrule
    \textbf{Total}                                   & \textbf{5742}                     & \textbf{5605}                       & \textbf{2763}                         & \textbf{1311} & \textbf{1294} & \textbf{1386} \\ \bottomrule \bottomrule
    \end{tabular}
    }
    \caption{Number of Examples by Category and Subcategory}
    \label{tab:data_size_extended}
    \end{table*}

\subsection{Overly Explanatory Examples} \label{app:overly_expl_examples}

    In Table~\ref{tab:overly_expl_examples} we present two examples where GPT-4 acting as an oracle translator (\S\ref{subsec:oracle_translation}) proposed overly explanatory translations in the emotional prosody category. Both are inline with the emotion of the speaker, but they contain new bits of information, not initially there. These were removed in filtering due to excessive word-length ratio between the two cases.

    \begin{table}[h]
        \resizebox{\columnwidth}{!}{
        \centering
        \begin{tabular}{@{}lll@{}}
        \midrule \midrule
        \multicolumn{3}{c}{Example 1: \emph{This will only take a minute.}}                                                         \\ \midrule
        A & (neutral) & Das dauert nur eine Minute. \\
         & & (This will only take a minute.) \\
        B & (angry)   & \begin{tabular}[c]{@{}l@{}}Das dauert nur eine Minute,\\ also machen Sie keinen Aufstand.\end{tabular} \\
        & & \begin{tabular}[c]{@{}l@{}}(This will only take a minute \\ so don't make a fuzz about it.)\end{tabular} \\ \midrule \midrule
        \multicolumn{3}{c}{Example 2: \emph{Our case was dismissed.}}                                                               \\ \midrule
        A & (neutral) & Unser Fall wurde abgewiesen.                                                                           \\
        & & (Our case was dismissed.) \\
        B & (sad)     & \begin{tabular}[c]{@{}l@{}}Unser Fall wurde abgewiesen\\ und das macht mich fassungslos.\end{tabular}  \\ 
        & & \begin{tabular}[c]{@{}l@{}}(Our case was dismissed\\ which is just perplexing.)\end{tabular} \\
        \midrule \midrule
        \end{tabular}
        }
    \caption{Examples of overly explanatory translations proposed by GPT-4.}
    \label{tab:overly_expl_examples}
    \end{table}

\section{Evaluated Speech Translation Models} \label{app:st_models}

     Here we describe in more detail the model families and the specific versions used. We evaluated in total 31 \stt{} model variants. All models are available in the Transformers Huggingface Library~\cite{hf}. For inference we used the default generation parameters and a beam search of 5.

    \begin{enumerate}

    \item \textsc{SeamlessM4T}~\citeSeamless{} and its updated version v2~\citeSeamlessv{} is a recently proposed family of unified encoder-decoder models that are both multilingual (many-to-many, 100 languages) and multimodal (speech/text input or output), meaning they can carry out the tasks of ASR, TTS, MT, S2TT, and also S2ST. The architecture is composed of a text encoder, text decoder, speech encoder, and speech decoder, and different parts are active depending on the input/output modalities. The text encoder-decoder is based on NLLB~\citenllb{}, the speech encoder on a newly proposed conformer~\cite{conformer} \textsc{w2v-BERT}~\cite{w2v_bert}, and the speech decoder on a unit decoder~\cite{unit_decoder} and a HiFi-GAN vocoder~\cite{vocoder}. The original version has a medium (1.2B)\footnote{\href{https://huggingface.co/facebook/seamless-m4t-medium}{hf.co/facebook/seamless-m4t-medium}} and a large (2.3B)\footnote{\href{https://huggingface.co/facebook/seamless-m4t-large}{hf.co/facebook/seamless-m4t-large}} variant, while the updated v2 has a large variant (2.3B)\footnote{\href{https://huggingface.co/facebook/seamless-m4t-v2-large}{hf.co/facebook/seamless-m4t-v2-large}}. For cascade \stt{} we first use the model in ASR mode, and then the same model is MT mode.
    
    \item \textsc{XLS-R}~\cite{xls-r} is a multilingual E2E \stt{} model that is based on a multilingual \textsc{wav2vec 2.0}~\cite{wav2vec2} trained with self-supervised learning on a large speech corpus on 128 languages. For \stt{}, the encoder is coupled with the decoder from \textsc{mBART50}~\cite{mbart}, and finetuned on paired speech-translation data. We use the folowing versions that are finetuned on English-to-15 on CoVoST2~\cite{covost2}: 300M\footnote{\href{https://huggingface.co/facebook/wav2vec2-xls-r-300m-en-to-15}{hf.co/facebook/wav2vec2-xls-r-300m-en-to-15}}, 1B\footnote{\href{https://huggingface.co/facebook/wav2vec2-xls-r-1b-en-to-15}{hf.co/facebook/wav2vec2-xls-r-1b-en-to-15}}, and 2B\footnote{\href{https://huggingface.co/facebook/wav2vec2-xls-r-2b-en-to-15}{hf.co/facebook/wav2vec2-xls-r-2b-en-to-15}}.

    \item \textsc{ZeroSwot} is a zero-shot E2E \stt{} model that softly connects a \textsc{wav2vec 2.0} encoder and an NLLB model, by compressing the speech representation into subword units and Optimal Transport~\cite{ot} alignment, using only ASR data. The versions used here are based on NLLB that were finetuned on the text data of CoVoST2, and the \textsc{ZeroSwot} model was trained on CommonVoice~\cite{common_voice}. The \textsc{Medium} version\footnote{\href{https://huggingface.co/johntsi/ZeroSwot-Medium_asr-cv_mt-covost2_en-to-15}{hf.co/johntsi/ZeroSwot-Medium-cv-covost2-en-to-15}} has 1B parameters and the \textsc{Large} version\footnote{\href{https://huggingface.co/johntsi/ZeroSwot-Large_asr-cv_mt-covost2_en-to-15}{hf.co/johntsi/ZeroSwot-Large-cv-covost2-en-to-15}} has 1.7B parameters.

    \item SALMONN~\cite{salmonn} is a general-purpose audio LLM that is capable of several speech- and audio-related tasks, including \stt{}. It is build on top of the Vicuna LLM~\cite{vicuna}, and uses two encoders, one from \textsc{Whisper} and one from BEATs~\cite{beats}. The concatenated output representations from the two encoders are processed by a Q-former~\cite{qformer} and fed to the LLM which is finetuned with LoRA~\cite{lora}. There is a 7B version\footnote{\href{https://huggingface.co/tsinghua-ee/SALMONN-7B}{hf.co/tsinghua-ee/SALMONN-7B}} and a 13B version\footnote{\href{https://huggingface.co/tsinghua-ee/SALMONN}{hf.co/tsinghua-ee/SALMONN}}. To translate speech into a target language we use the recommended prompt from the paper: ``Listen to the speech and translate it into \{Target Language\}''.

    \item \textsc{Whisper} \& NLLB is an AED-based cascade. \textsc{Whisper}~\cite{whisper} is an encoder-decoder ASR and many-to-en \stt{} model. We use three different versions for this casdade, namely the \textsc{Whisper-Medium}\footnote{\href{https://huggingface.co/openai/whisper-medium}{hf.co/openai/whisper-medium}}, the \textsc{Whisper-Large}\footnote{\href{https://huggingface.co/openai/whisper-large}{hf.co/openai/whisper-large}}, and the latest v3 large version\footnote{\href{https://huggingface.co/openai/whisper-large-v3}{hf.co/openai/whisper-large-v3}}. We primarily present results with the \textsc{Whisper-Large-v3}, but since it was also used for filtering we also discuss v1 in order to avoid biasing our results. NLLB~\citenllb{} is a massively multilingual many-to-many MT model with access to 200 languages. We used the two distilled versions from the 54B MoE model, namely the distilled-600M\footnote{\href{https://huggingface.co/facebook/nllb-200-distilled-600M}{hf.co/facebook/nllb-200-distilled-600M}} and the distilled-1.3B\footnote{\href{https://huggingface.co/facebook/nllb-200-distilled-1.3B}{hf.co/facebook/nllb-200-distilled-1.3B}}, as well as the 3.3B model\footnote{\href{https://huggingface.co/facebook/nllb-200-3.3B}{hf.co/facebook/nllb-200-3.3B}}. We evaluated all possible combinations, thus having 9 cascade variants with these models.

    \item CTC \& NLLB is a CTC-based cascade. We use three different CTC encoders for the cascades. The first one is the Large version (300M) of \textsc{wav2vec 2.0}\footnote{\href{https://huggingface.co/facebook/wav2vec2-large-960h-lv60-self}{hf.co/facebook/wav2vec2-large-960h-lv60-self}} which is finetuned on Libri-Light~\cite{librilight} and Librispeech~\cite{librispeech}, additionally using self-training~\cite{self-training}. The second is the Large version (300M) of \textsc{HuBERT}\footnote{\href{https://huggingface.co/facebook/hubert-large-ls960-ft}{hf.co/facebook/hubert-large-ls960-ft}}~\cite{hubert}, finetuned on Librispeech. The third is also based on \textsc{HuBERT}, more specifically to the XL version\footnote{\href{https://huggingface.co/facebook/hubert-xlarge-ls960-ft}{hf.co/facebook/hubert-xlarge-ls960-ft}} with 1B parameters. We use the same three versions of NLLB, as we did for the AED-based cascade, thus having in total 9 variants of the CTC-based cascade.
    
    \end{enumerate}

\section{Supplementary Results} \label{app:results}

\begin{figure}[h]
        \centering
        \includegraphics[width=\columnwidth]{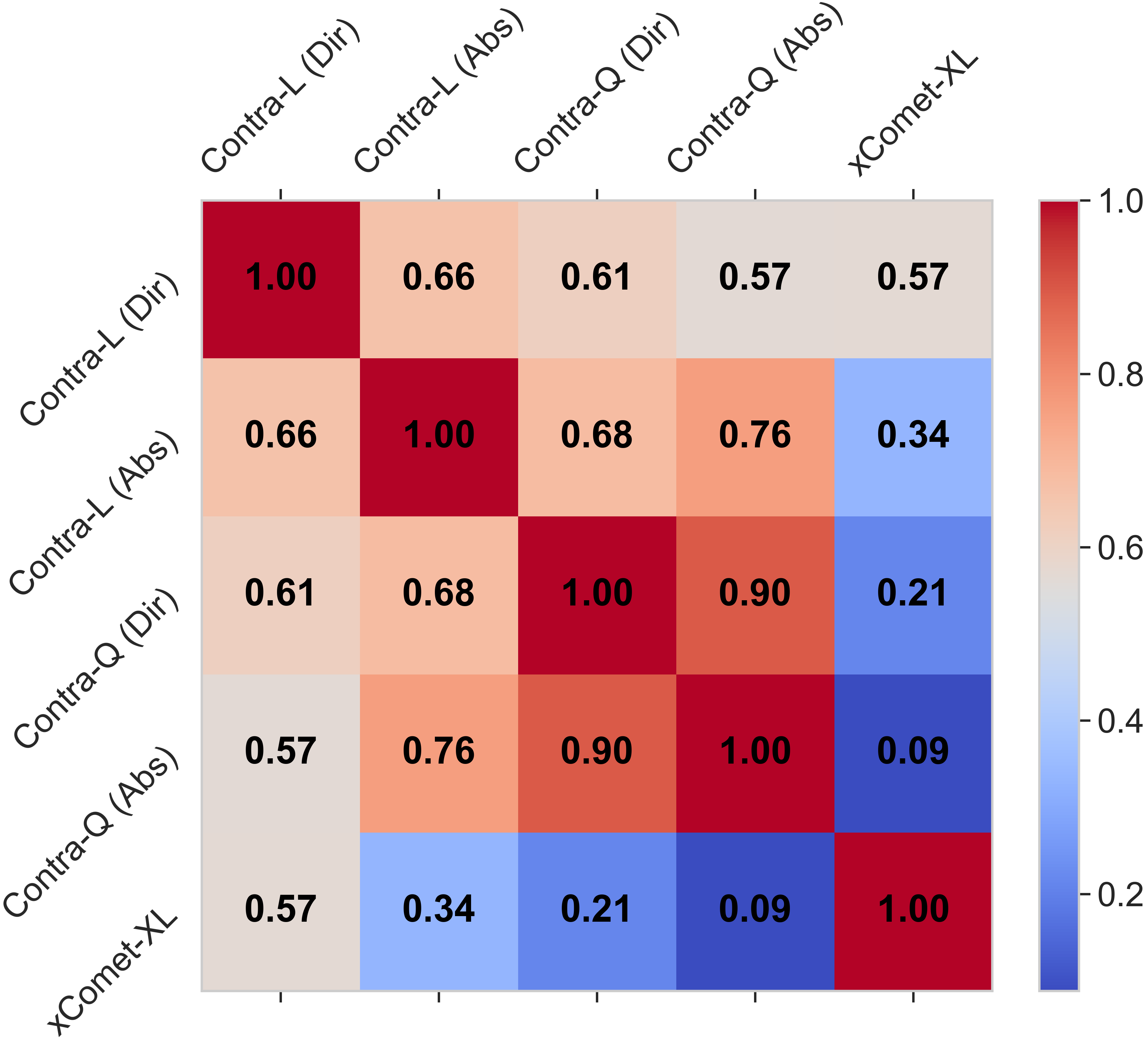}
        \caption{Correlation Matrix of the metrics across all language pairs and models.}
        \label{fig:metric_correlation}
    \end{figure}

    In Figure~\ref{fig:metric_correlation} we present the Spearman rank correlation for the four contrastive metrics and the standard evaluation metric \textsc{xCOMET}. They were computed by evaluating all 31 models (\S\ref{app:st_models}) for all 3 language pairs, thus having a total of 93 observations. 

    In Table~\ref{tab:contra_q_scores_all} we present the global contrastive quality scores for all 31 \stt{} models for the 3 language pairs, which were used for the analysis of Figure~\ref{fig:regression_strong} in \S\ref{sec:results} of the main text.
    
    \begin{table*}[h]
    \centering
    \resizebox{0.95\textwidth}{!}{
    \begin{tabular}{@{}lcccccc}
    \toprule
    \textbf{Model} & \textbf{Model Type} & \textbf{Model Size (B)} & \multicolumn{4}{c}{\textbf{Contrastive Quality (Global)}} \\ 
    \cmidrule(lr){4-7}
    & & & \textbf{En-De} & \textbf{En-Es} & \textbf{En-Ja} & \textbf{Average} \\ \midrule
    \textsc{SeamlessM4T-v1-Medium}         & E2E         & 1.2 & 9.1 & 10.4 & 10.1 & 9.9 \\
    \textsc{SeamlessM4T-v1-Large}          & E2E         & 2.3 & 7.8 & 8.7 & 8.2 & 8.2 \\
    \textsc{SeamlessM4T-v2-Large}          & E2E         & 2.3 & 14.5 & 11.0 & 13.9 & 13.1 \\ \midrule
    XLS-R 300M                    & E2E         & 0.3 & 10.2 & 9.3 & 9.5 & 9.6 \\
    XLS-R 1B                      & E2E         & 1.0 & 9.3 & 8.3 & 8.1 & 8.6 \\
    XLS-R 2B                      & E2E         & 2.0 & 7.3 & 8.4 & 7.0 & 7.6 \\ \midrule
    \textsc{ZeroSwot-Medium}              & E2E      & 0.9 & 8.9 & 7.5 & 7.5 & 8.0 \\
    \textsc{ZeroSwot-Large}                & E2E      & 0.9 & 8.7 & 7.7 & 7.9 & 8.1 \\ \midrule
    SALMONN-7B                    & E2E     & 7.0 & 12.7 & 9.4 & 12.9 & 11.7 \\
    SALMONN-13B                   & E2E     & 13.0 & \textbf{15.9} & \textbf{12.3} & \textbf{16.1} & \textbf{14.8} \\ \midrule
    \textsc{SeamlessM4T-v1-Medium}         & Cascade-AED & 2.4 & 7.3 & 7.6 & 6.9 & 7.2 \\
    \textsc{SeamlessM4T-v1-Large}          & Cascade-AED & 4.6 & 6.5 & 5.8 & 4.3 & 5.5 \\
    \textsc{SeamlessM4T-v2-Large}          & Cascade-AED & 4.6 & 10.5 & 7.7 & 8.7 & 8.9 \\ \midrule
    \textsc{Whisper-v1-Medium} \& NLLB-600M & Cascade-AED & 1.4 & 5.7 & 5.0 & 5.4 & 5.4 \\
    \textsc{Whisper-v1-Medium} \& NLLB-1.3B & Cascade-AED & 2.1 & 6.2 & 5.0 & 5.8 & 5.6 \\
    \textsc{Whisper-v1-Medium} \& NLLB-3.3B & Cascade-AED & 4.1 & 6.6 & 5.1 & 5.4 & 5.7 \\
    \textsc{Whisper-v1-Large} \& NLLB-600M  & Cascade-AED & 2.2 & 5.9 & 4.9 & 6.1 & 5.6 \\
    \textsc{Whisper-v1-Large} \& NLLB-1.3B  & Cascade-AED & 2.9 & 6.0 & 4.7 & 5.8 & 5.5 \\
    \textsc{Whisper-v1-Large} \& NLLB-3.3B  & Cascade-AED & 4.9 & 6.3 & 4.6 & 5.6 & 5.5 \\
    \textsc{Whisper-v3-Large} \& NLLB-600M  & Cascade-AED & 2.2 & 5.3 & 4.6 & 6.5 & 5.5 \\
    \textsc{Whisper-v3-Large} \& NLLB-1.3B  & Cascade-AED & 2.9 & 5.3 & 4.7 & 5.4 & 5.1 \\
    \textsc{Whisper-v3-Large} \& NLLB-3.3B  & Cascade-AED & 4.9 & 5.5 & 4.8 & 5.3 & 5.2 \\ \midrule
    \textsc{wav2vec 2.0} \& NLLB-600M       & Cascade-CTC & 0.9 & 1.4 & 1.3 & 2.0 & 1.5 \\
    \textsc{wav2vec 2.0} \& NLLB-1.3B       & Cascade-CTC & 1.6 & 1.7 & 1.0 & 1.4 & 1.3 \\
    \textsc{wav2vec 2.0} \& NLLB-3.3B       & Cascade-CTC & 3.6 & 1.6 & 0.9 & 1.7 & 1.4 \\
    \textsc{HuBERT} \& NLLB-600M            & Cascade-CTC & 0.9 & 3.2 & 2.7 & 2.5 & 2.8 \\
    \textsc{HuBERT} \& NLLB-1.3B            & Cascade-CTC & 1.6 & 2.2 & 2.4 & 2.9 & 2.5 \\
    \textsc{HuBERT} \& NLLB-3.3B            & Cascade-CTC & 3.6 & 2.7 & 2.6 & 2.9 & 2.7 \\
    \textsc{HuBERT-xl} \& NLLB-600M         & Cascade-CTC & 1.6 & 2.4 & 2.9 & 3.0 & 2.8 \\
    \textsc{HuBERT-xl} \& NLLB-1.3B         & Cascade-CTC & 2.3 & 3.5 & 1.7 & 2.7 & 2.6 \\
    \textsc{HuBERT-xl} \& NLLB-3.3B         & Cascade-CTC & 4.3 & 2.6 & 2.4 & 2.5 & 2.5 \\ \bottomrule
    \end{tabular}
    }
    \caption{Contrastive Quality (Global) scores for English-German, English-Spanish, and English-Japanese, including their averages.}
    \label{tab:contra_q_scores_all}
    \end{table*}

\begin{figure*}[h]
        \centering
        \begin{minipage}[t]{0.8\textwidth}
            \centering
            \includegraphics[width=\textwidth]{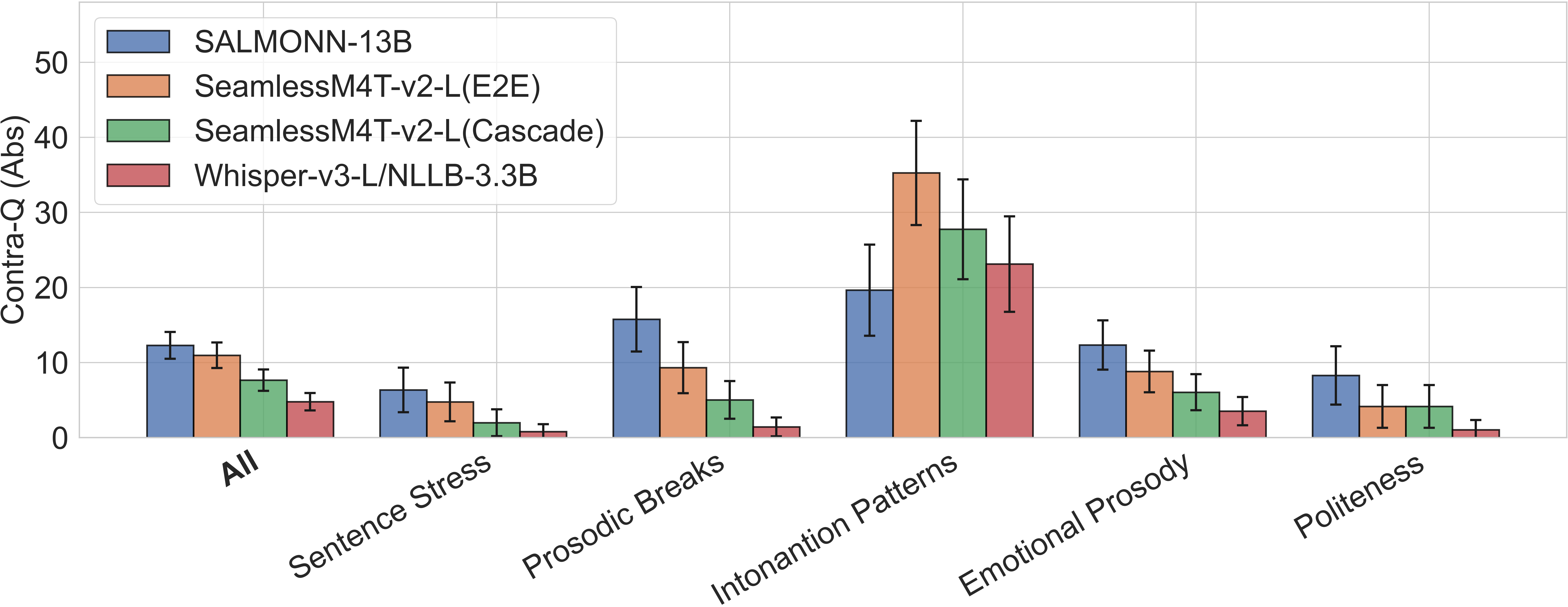}
        \end{minipage}
        \hfill
        \begin{minipage}[t]{\textwidth}
            \centering
            \includegraphics[width=0.8\textwidth]{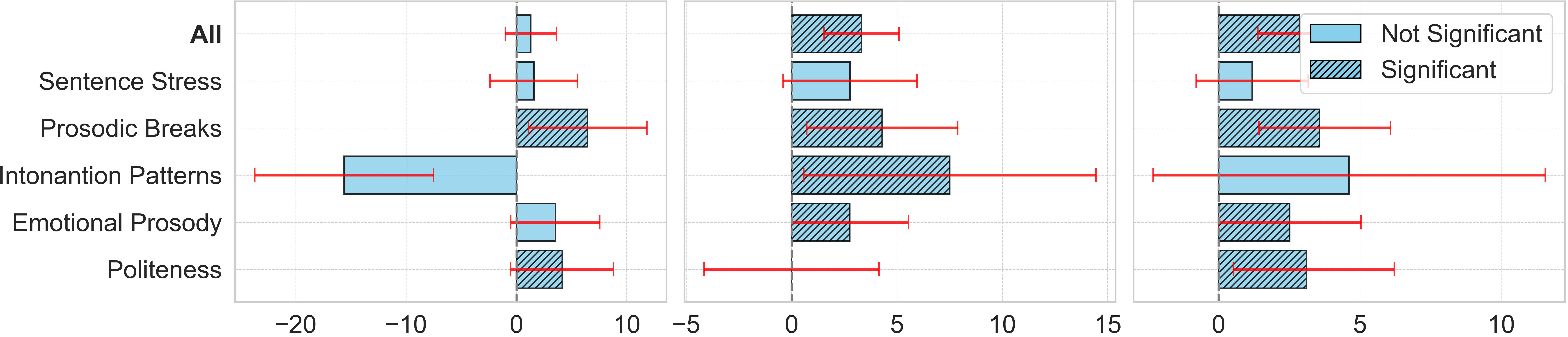}
        \end{minipage}
        \caption{Upper: Model performance per category (\texttt{En}-\texttt{Es}). Lower: Model performance comparisons (\texttt{En}-\texttt{Es}), (a): SALMONN-13B vs.\ \textsc{SeamlessM4T-v2-Large}, (b) \textsc{SeamlessM4T-v2-Large}(E2E) vs.\ \textsc{SeamlessM4T-v2-Large}(cascade), (c) \textsc{SeamlessM4T-v2-Large}(cascade) vs.\ \textsc{Whisper-v3-Large}/NLLB-3.3B.}
        \label{fig:performance_comparisons_spa}
    \end{figure*}

\begin{figure*}[h]
        \centering
        \begin{minipage}[t]{0.8\textwidth}
            \centering
            \includegraphics[width=\textwidth]{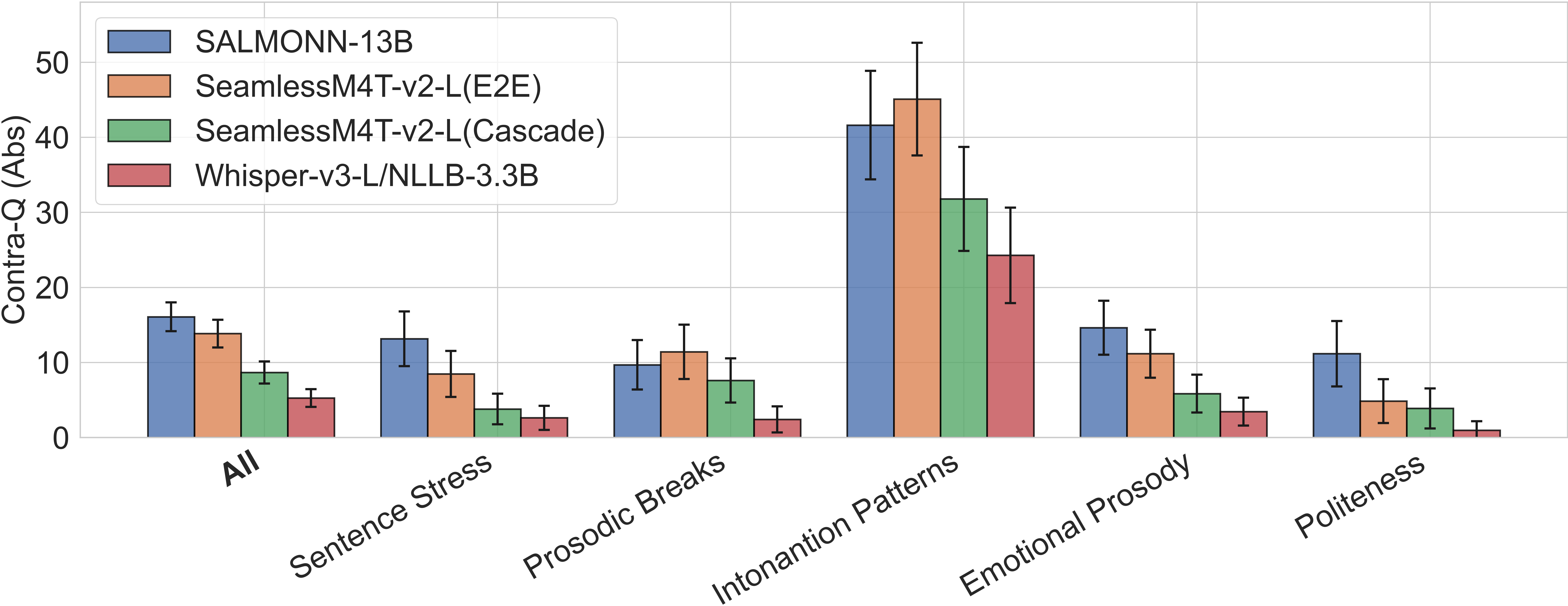}
        \end{minipage}
        \hfill
        \begin{minipage}[t]{\textwidth}
            \centering
            \includegraphics[width=0.8\textwidth]{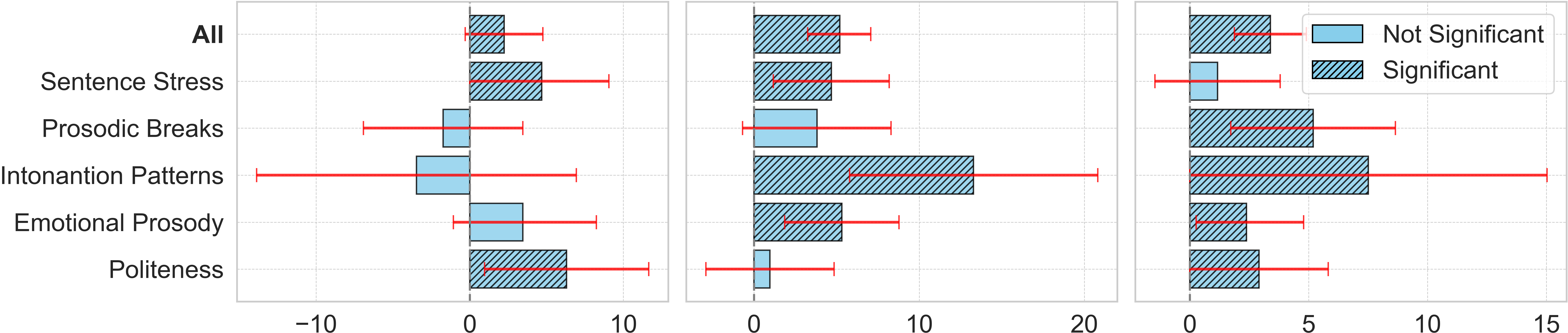}
        \end{minipage}
        \caption{Upper: Model performance per category (\texttt{En}-\texttt{Ja}). Lower: Model performance comparisons (\texttt{En}-\texttt{Ja}), (a): SALMONN-13B vs.\ \textsc{SeamlessM4T-v2-Large}, (b) \textsc{SeamlessM4T-v2-Large}(E2E) vs.\ \textsc{SeamlessM4T-v2-Large}(cascade), (c) \textsc{SeamlessM4T-v2-Large}(cascade) vs.\ \textsc{Whisper-v3-Large}/NLLB-3.3B.}
        \label{fig:performance_comparisons_jpn}
    \end{figure*}

    In Figures~\ref{fig:performance_comparisons_spa} and \ref{fig:performance_comparisons_jpn} we present the comparisons of the 4 models for Spanish and Japanese, similar to what we did in Figure~\ref{fig:performance_comparisons_deu} for German in the main text. In general, the findings and observations here coincide with those for German.

\end{document}